\documentclass[sigconf]{acmart}
\usepackage{placeins}
\usepackage{booktabs} 
\usepackage{bm}
\usepackage{mathtools}
\usepackage{subcaption}
\usepackage{algorithm}
\usepackage{algorithmic}
\usepackage{color}
\usepackage[utf8]{inputenc}

\DeclareUnicodeCharacter{2041}{}

\setcopyright{acmlicensed}

\acmDOI{10.1145/3321707.3321721}

\acmISBN{978-1-4503-6111-8/19/07}

\acmConference[GECCO '19]{the Genetic and Evolutionary Computation Conference 2019}{July 13--17, 2019}{Prague, Czech Republic}
\acmYear{2019}
\copyrightyear{2019}

\acmPrice{15.00}

\acmBooktitle{Genetic and Evolutionary Computation Conference (GECCO '19), July 13--17, 2019, Prague, Czech Republic}

\begin{document}
\title{Evolutionary Neural AutoML for Deep Learning}

\author{Jason Liang, Elliot Meyerson, Babak Hodjat, Dan Fink, Karl Mutch, and Risto Miikkulainen}
\authornote{firstname.lastname@cognizant.com}
\affiliation{%
  \institution{Cognizant Technology Solutions}
}
\affiliation{%
  \institution{The University of Texas at Austin}
}

\renewcommand{\shortauthors}{Jason Liang et al.}

\begin{abstract}
Deep neural networks (DNNs) have produced state-of-the-art results in many benchmarks and problem domains.
However, the success of DNNs depends on the proper configuration of its architecture and hyperparameters.
Such a configuration is difficult and as a result, DNNs are often not used to their full potential.
In addition, DNNs in commercial applications often need to satisfy real-world design constraints such as size or number of parameters.
To make configuration easier, automatic machine learning (AutoML) systems for deep learning have been developed, focusing mostly on optimization of hyperparameters.

This paper takes AutoML a step further.
It introduces an evolutionary AutoML framework called LEAF that not only optimizes hyperparameters but also network architectures and the size of the network.
LEAF makes use of both state-of-the-art evolutionary algorithms (EAs) and distributed computing frameworks.
Experimental results on medical image classification and natural language analysis show that the framework can be used to achieve state-of-the-art performance.
In particular, LEAF demonstrates that architecture optimization provides a significant boost over hyperparameter optimization, and that networks can be minimized at the same time with little drop in performance.
LEAF therefore forms a foundation for democratizing and improving AI, as well as making AI practical in future applications.

\end{abstract}

%
%


\maketitle

\section{Introduction}\label{intro}

Applications of machine learning and artificial intelligence have increased significantly recently, driven by both improvements in computing power and quality of data.
In particular, deep neural networks (DNN) \cite{lecun2015deep} learn rich representations of high-dimensional data, exceeding the state-of-the-art in an variety of benchmarks in computer vision, natural language processing, reinforcement learning, and speech recognition \cite{collobert:icml08,graves:icassp13,he:arxiv16}.
Such state-of-the-art DNNs are very large, consisting of hundreds of millions of parameters, requiring large computational resources to train and run.
They are also highly complex, and their performance depends on their architecture and choice of hyperparameters \cite{he:arxiv16,ng:arxiv15,che:arxiv16}.

Much of the recent research in deep learning indeed focuses on discovering specialized architectures that excel in specific tasks.
There is much variation between DNN architectures (even for single-task domains) and so far, there are no guiding principles for deciding between them.
Finding the right architecture and hyperparameters is essentially reduced to a black-box optimization process.
However, manual testing and evaluation is a tedious and time consuming process that requires experience and expertise.
The architecture and hyperparameters are often chosen based on history and convenience rather than theoretical or empirical principles, and as a result, the network has does not perform as well as it could.
Therefore, automated configuration of DNNs is a compelling approach for three reasons: (1) to find innovative configurations of DNNs that also perform well, (2) to find configurations that are small enough to be practical, and (3) to make it possible to find them without domain expertise.

Currently, the most common approach to satisfy the first goal is through partial optimization.
The authors might tune a few hyperparameters or switch between several fixed architectures, but rarely optimize both the architecture and hyperparameters simultaneously \cite{vincent2010stacked, he2016identity}.
This approach is understandable since the search space is massive and existing methods do not scale as the number of hyperparameters and architecture complexity increases.
The standard and most widely used methods for hyperparameter optimization is grid search, where hyperparameters are discretized into a fixed number of intervals and all combinations are searched exhaustively.
Each combination is tested by training a DNN with those hyperparameters and evaluating its performance with respect to a metric on a benchmark dataset.
While this method is simple and can be parallelized easily, its computational complexity grows combinatorially with the number of hyperparameters, and becomes intractable once the number of hyperparameters exceeds four or five \cite{keogh2011curse}.
Grid search also does not address the question of what the optimal architecture of the DNN should be, which may be just as important as the choice of hyperparameters.
A method that can optimize both structure and parameters is needed.

Recently, commercial applications of deep learning have become increasingly important and many of them run on smartphones.
Unfortunately, the hundreds of millions of weights of modern DNNs cannot fit to the few gigabytes of RAM in most smartphones.
Therefore, an important second goal of DNN optimization is to minimize the complexity or size of a network, while simultaneously maximizing its performance \cite{howard2017mobilenets}.
Thus, a method for optimizing multiple objectives is needed to meet the second goal.

In order to achieve the third goal, i.e. democratizing AI, systems for automating DNN configuration have been developed, such as Google AutoML \cite{googleaiblog_2017} and Yelp's Metric Optimization Engine (MOE \cite{moe}, also commercialized as a product called SigOpt \cite{sigopt}).
However, existing systems are often limited in both the scope of the problems they solve and how much feedback they provide to the user.
For example, Google AutoML system is a black-box that hides the network architecture and training from the user; it only provides an API by which the user can use to query on new inputs.
MOE is more transparent on the other hand, but since it uses a Bayesian optimization algorithm underneath, it only tunes hyperparameters of a DNN.
Neither systems minimizes the size or complexity of the networks.

The main contribution of this paper is a novel AutoML system called LEAF (Learning Evolutionary AI Framework) that addresses these three goals.
LEAF leverages and extends an existing state-of-the-art evolutionary algorithm for architecture search called CoDeepNEAT \cite{miikkulainen2017evolving}, which evolves both hyperparameters and network structure.
While its hyperparameter optimization ability already matches those of other AutoML systems, LEAF’s additional ability to optimize DNN architectures further makes it possible to achieve state-of-the-art results.
The speciation and complexification heuristics inside CoDeepNEAT also allows it to be easily adapted to multiobjective optimization to find minimal architectures.
The effectiveness of LEAF will be demonstrated in this paper on two domains, one in language: Wikipedia comment toxicity classification (also referred to as Wikidetox), and another in vision: Chest X-rays multitask image classification.
LEAF therefore forms a foundation for democratizing, simplifying, and improving AI.

\section{Background and Related Work}

This section will review background and related work in hyperparameter optimization and neural architecture search.

\subsection{Hyperparameter Tuning for DNNs}

As mentioned in Section~\ref{intro}, the simplest form of hyperparameter optimization is exhaustive grid search, where points in hyperparameter space are sampled uniformly at regular intervals \cite{vincent2010stacked}.
A straightforward extension is random search, where the points are sampled uniformly at random from the search space \cite{bergstra2012random}.
These methods can optimize simple DNNs, but are ineffective when all hyperparameters are crucial to performance and must be tuned to very particular values.
For networks with such characteristics, Bayesian optimization using Gaussian processes \cite{snoek2012practical} is a feasible alternative.
Bayesian optimization requires relatively few function evaluations and works well on multimodal, non-separable, and noisy functions where there are several local optima.
It first creates a probability distribution of functions (also known as Gaussian process) that best fits the objective function and then uses that distribution to determine where to sample next.
The main weakness of Bayesian optimization is that it is computational expensive and scales cubically with the number of evaluated points. DNGO \cite{snoek2015scalable} tried to address this issue by replacing Gaussian processes with linearly scaling Bayesian neural networks.
Another downside of Bayesian optimization it performs poorly when the number of hyperparameters is moderately high, i.e. more than 10-15 \cite{loshchilov2016cma}.

EAs are another class of algorithms widely used for black-box optimization of complex, multimodal functions.
They rely on biological inspired mechanisms to improve iteratively upon a population of candidate solutions to the objective function.
One particular EA that has been successfully applied to DNN hyperparameter tuning is CMA-ES \cite{loshchilov2016cma}.
In CMA-ES, a Gaussian distribution for the best individuals in the population is estimated and used to generate/sample the population for the next generation.
Furthermore, it has mechanisms for controlling the step-size and the direction that the population will move.
CMA-ES has been shown to perform well in many real-world high-dimensional optimization problems and in particular, CMA-ES has been shown to outperform Bayesian optimization on tuning the parameters of a convolutional neural network \cite{loshchilov2016cma}.
It is however limited to continuous optimization and there does not extend naturally to architecture search.

\subsection{Architecture Search for DNNs}

One recent approach is to use reinforcement learning (RL) to search for better architectures.
A recurrent neural network (LSTM) controller generates a sequence of layers that begin from the input and end at the output of a DNN \cite{zoph:arxiv16}.
The LSTM is trained through a gradient-based policy search algorithm called REINFORCE \cite{williams1992simple}.
The architecture search space explored by this approach is sufficiently large to improve upon hand-design.
On popular image classification benchmarks such as CIFAR-10 and ImageNet, such an approach achieved performance within 1-2 percentage points of the state-of-the-art, and on a language modeling benchmark, it achieved state-of-the-art performance at the time \cite{zoph:arxiv16}. 

However, the architecture of the optimized network still must have either a linear or tree-like core structure; arbitrary graph topologies are outside the search space.
Thus, it is still up to the user to define an appropriate search space beforehand for the algorithm to use as a starting point.
The number of hyperparameters that can be optimized for each layer are also limited.
Furthermore, the computations are extremely heavy; to generate the final best network, many thousands of candidate architectures have to be evaluated and trained, which requires hundreds of thousands of GPU hours.

An alternative direction for architecture search is evolutionary algorithms (EAs).
They are well suited for this problem because they are black-box optimization algorithms that can optimize arbitrary structure.
Some of these approaches use a modified version of NEAT \cite{real2017large}, an EA for neuron-level neuroevolution \cite{stanley:ec02}, for searching network topologies.
Others rely on genetic programming \cite{suganuma2017genetic} or hierarchical evolution \cite{liu2017hierarchical}.
There is some very recent work on multiobjective evolutionary architecture search \cite{elsken1804efficient,lu2018nsga}, where the goal is to optimize both the performance and training time/complexity of the network.

The main advantage of EAs over RL methods is that they can optimize over much larger search spaces.
For instance, approaches based on NEAT \cite{real2017large} can evolve arbitrary graph topologies for the network architecture.
Most importantly, hierarchical evolutionary methods \cite{liu2017hierarchical}, can search over very large spaces efficiently and evolve complex architectures quickly from a minimal starting point.
As a result, the performance of evolutionary approaches match or exceed that of reinforcement learning methods.
For example, the current state-of-the-art results on CIFAR-10 and ImageNet were achieved by an evolutionary approach \cite{real2018regularized}.
In this paper, LEAF uses CoDeepNEAT, a powerful EA based on NEAT that is capable of hierarchically evolving networks with arbitrary topology.

\section{LEAF Overview}

\begin{figure}[t]
 \begin{center}
   \includegraphics[width=\linewidth]{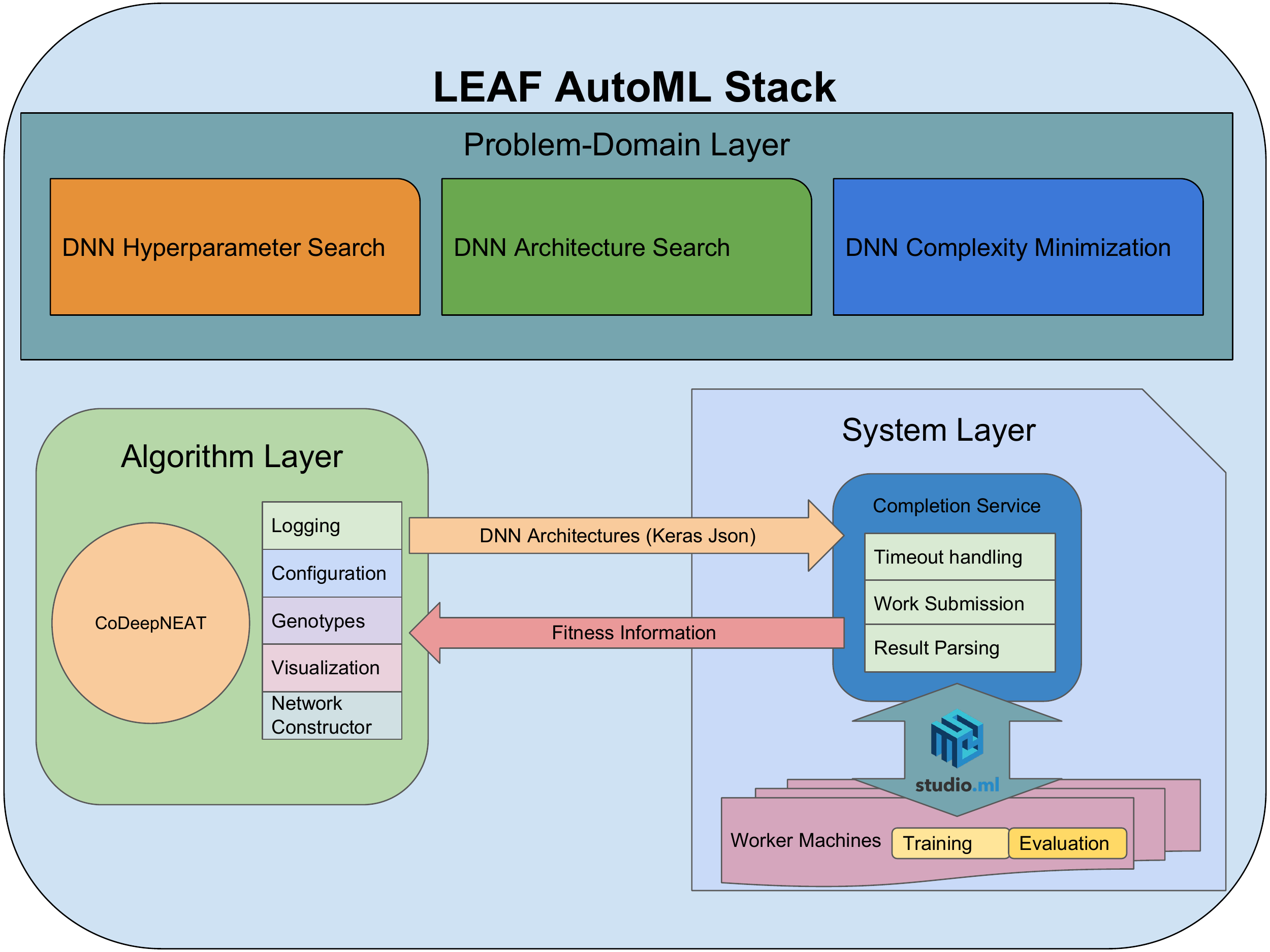}
   \caption{A visualization of LEAF and its internal subsystems. The three main components are: (1) the algorithm layer which uses CoDeepNEAT to evolve hyperparameters or neural networks, (2) the system layer which helps train and evaluate the networks evolved by the algorithm layer, and (3) the problem-domain layer, which utilizes the two previous layers to optimize DNNs. The decoupling of the algorithm and system layers allows LEAF to be easily applied to varying problem types, e.g., via options for multiobjective optimization and different types of neural network layers.}
   \label{fig:leaf}
 \end{center}
\end{figure}

LEAF is an AutoML system composed of three main components: algorithm layer, system layer, and problem-domain layer.
The algorithm layer allows the LEAF to evolve DNN hyperparameters and architectures.
The system layer parallelizes training of DNNs on cloud compute infrastructure such as Amazon AWS \cite{amazon}, Microsoft Azure \cite{azure}, or Google Cloud \cite{gcloud}, which is required to evaluate the fitnesses of the networks evolved in the algorithm layer.
The algorithm layer sends the network architectures in Keras JSON format \cite{Chollet:2015} to the system layer and receives fitness information back.
These two layers work in tandem to support the problem-domain layer, where LEAF solves problems such as hyperparameter tuning, architecture search, and complexity minimization.
An overview of LEAF AutomML's structure is shown in Figure~\ref{fig:leaf}.

\subsection{Algorithm Layer}

\begin{figure}[t]
  \begin{center}
    \includegraphics[width=\linewidth]{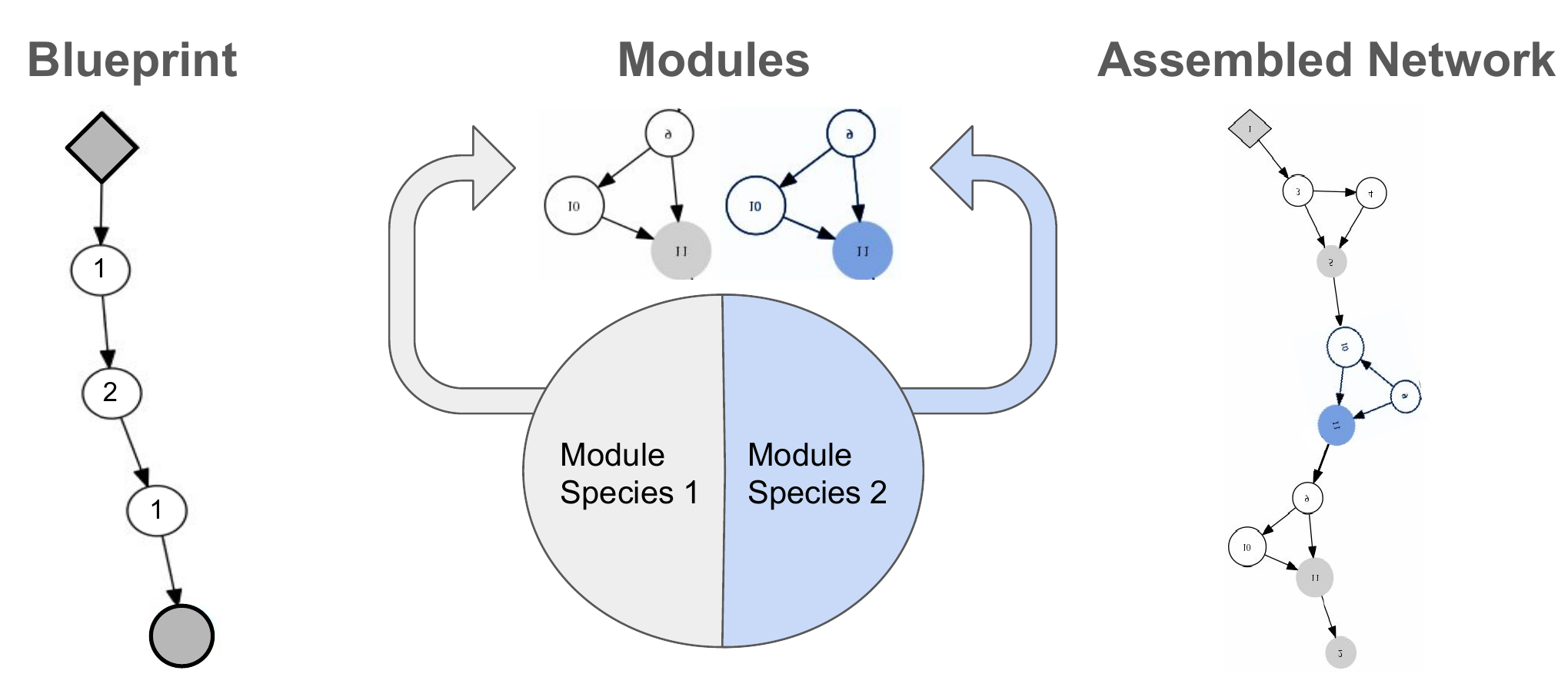}
    \caption{A visualization of how CoDeepNEAT assembles networks for       fitness evaluation. Modules and blueprints are assembled together into a network through replacement of blueprint nodes with corresponding modules. This approach allows evolving repetitive and deep structures seen in many hand-designed DNNs.}
    \label{fig:codeepneat_assembling}
  \end{center}
\end{figure}

The core of the algorithm layer is composed of CoDeepNEAT, an cooperative coevolutionary algorithm based on NEAT for evolving DNN architectures and hyperparameters \cite{miikkulainen2017evolving}.
Cooperative coevolution is a commonly used technique in evolutionary computation to discover complex behavior during evaluation by combining simpler components together.
It has been used with success in many domains, including function optimization \cite{potter1994cooperative}, predator-prey dynamics \cite{yong2001cooperative}, and subroutine optimization \cite{yanai2001multi}.
The specific coevolutionary mechanism in CoDeepNEAT is inspired by Hierarchical SANE \cite{moriarty1998hierarchical} but is also influenced by component-evolution approaches of ESP \cite{gomez1999solving} and CoSyNE \cite{gomez2008accelerated}.
These methods differ from conventional neuroevolution in that they do not evolve entire networks.
Instead, both approaches evolve components that are then assembled into complete networks for fitness evaluation.

CoDeepNEAT follows the same fundamental process as NEAT:
First, a population of chromosomes of minimal complexity is created.
Each chromosome is represented as a graph and is also referred to as an individual.
Over generations, structure (i.e.\ nodes and edges) is added to the graph incrementally through mutation.
As in NEAT, mutation involves randomly adding a node or a connection between two nodes.
During crossover, historical markings are used to determine how genes of two chromosomes can be lined up and how nodes can be randomly crossed over.
The population is divided into species (i.e.\ subpopulations) based on a similarity metric.
Each species grows proportionally to its fitness and evolution occurs separately in each species.

CoDeepNEAT differs from NEAT in that each node in the chromosome no longer represents a neuron, but instead a layer in a DNN.
Each node contains a table of real and binary valued hyperparameters that are mutated through uniform Gaussian distribution and random bit-flipping, respectively.
These hyperparameters determine the type of layer (such as convolutional, fully connected, or recurrent) and the properties of that layer (such as number of neurons, kernel size, and activation function).
The edges in the chromosome are no longer marked with weights; instead they simply indicate how the nodes (layers) are connected.
The chromosome also contains a set of global hyperparameters applicable to the entire network (such as learning rate, training algorithm, and data preprocessing).

  \begin{algorithm}[t]
  \caption{CoDeepNEAT}\label{alg:codeepneat}
  \flushleft
  \textbf{1. Given} population of modules/blueprints\\
  \textbf{2. For each} blueprint $B_i$ during every generation:\\
      \hspace{10pt} \textbf{3. For each} node $N_j$ in $B_i$\\
      \hspace{20pt} \textbf{4. Choose} randomly from module species that $N_j$ points to\\
      \hspace{20pt} \textbf{5. Replace} $N_j$ with randomly chosen module $M_j$\\
      \hspace{10pt} \textbf{6. When} all nodes in $B_i$ are replaced, convert $B_i$ to assembled network $N_i$\\
  \textbf{7. Evaluate} fitnesses of the assembled networks $N$\\
  \textbf{8. For each} network $N_i$\\
      \hspace{10pt} \textbf{9. Attribute} fitness of $N_i$ to its component blueprint $B_i$ and modules $M_j$\\
  \textbf{10. Evolve} blueprint and module population with NEAT\\
  \end{algorithm}

As summarized in Algorithm~\ref{alg:codeepneat}, two populations of modules and blueprints are evolved separately using mutation and crossover operators of NEAT.
The blueprint chromosome (also known as an individual) is a graph where each node contains a pointer to a particular module species.
In turn, each module chromosome is a graph that represents a small DNN.
During fitness evaluation, the modules and blueprints are combined to create a large assembled network.
For each blueprint chromosome, each node in the blueprint's graph is replaced with a module chosen randomly from the species to which that node points.
If multiple blueprint nodes point to the same module species, then the same module is used in all of them.
After the nodes in the blueprint have been replaced, the individual is converted into a DNN.
This entire process for assembling the network is visualized in Figure~\ref{fig:codeepneat_assembling}.

The assembled networks are evaluated by first letting the networks learn on a training dataset for the task and then measuring their performance with an unseen validation set.
The fitnesses, i.e. validation performance, of the assembled networks are attributed back to blueprints and modules as the average fitness of all the assembled networks containing that blueprint or module.
This scheme reduces evaluation noise and allows blueprints or modules to be preserved into the next generation even if they be occasionally included in a poorly performing network.
After CoDeepNEAT finishes running, the best evolved network is trained until convergence and evaluated on another holdout testing set.

\subsection{System Layer}

One of the main challenges in using CoDeepNEAT to evolve the architecture and hyperparameters of DNNs is the computational power required to evaluate the networks.
However, because evolution is a parallel search method, the evaluation of individuals in the population every generation can be distributed over hundreds of worker machines, each equipped with a dedicated GPU.
For most of the experiments described in this paper, the workers are GPU equipped machines running on Microsoft Azure, a popular platform for cloud computing \cite{azure}.

To this end, the system layer of LEAF uses the API called the completion service that is part of an open-source package called StudioML \cite{studioml}.
First, the algorithm layer sends networks ready for fitness evaluation in the form of Keras JSON to the system layer server node.
Next, the server node submits the networks to the completion service.
They are pushed onto a queue (buffer) and each available worker node pulls a  single network from the queue to train.
After training is finished, fitness is calculated for the network and the information is immediately returned to the server.
The results are returned one at a time and without any order guarantee through a separate return queue.
By using the completion service to parallelize evaluations, thousands of candidate networks are trained in a matter of days, thus making architecture search tractable.

\subsection{Problem-Domain Layer}

The problem-domain layer solves the three tasks mentioned earlier, i.e. optimization of hyperparameters, architecture, and network complexity, using CoDeepNEAT is a starting point.

\textit{Hyperparameter Optimization.}
By default, LEAF optimizes both architecture and hyperparameters.
To demonstrate the value of architecture search, it is possible to configure CoDeepNEAT in the algorithm layer to optimize hyperparameters only.
In this case, mutation and crossover of network structure and node-specific hyperparameters are disabled.
Only the global set of hyperparameters contained in each chromosome are optimized, as in the case in other hyperparameter optimization methods.
Hyperparameter-only CoDeepNEAT is very similar to a conventional genetic algorithm in that there is elitist selection and the hyperparameters undergo uniform mutation and crossover.
However, it still has NEAT's speciation mechanism, which protects new and innovative hyperparameters by grouping them into subpopulations.

\textit{Architecture Search.}
LEAF directly utilizes standard CoDeepNEAT to perform architecture search in simpler domains such as single-task classification.
However, LEAF can also be used to search for DNN architectures for multitask learning (MTL).
The foundation is the soft-ordering multitask architecture \cite{Meyerson:2018} where each level of a fixed linear DNN consists of a fixed number of modules.
These modules are then used to a different degree for the different tasks.
LEAF extends this MTL architecture by coevolving both the module architectures and the blueprint (routing between the modules) of the DNN \cite{Liang2018Evolutionary}.

\begin{algorithm}[t]
\caption{Multiobj CoDeepNEAT Module/Blueprint Ranking}\label{alg:mcdn_lower}
\flushleft
\textbf{1. Given} population of modules/blueprints, evaluated primary and secondary objectives ($X$ and $Y$)\\
\textbf{2. For each} species $S_i$ during every generation:\\
    \hspace{10pt} \textbf{3. Create} new empty species $\hat{S}_i$\\
    \hspace{10pt} \textbf{4. While} $S_i$ is not empty\\
    \hspace{20pt} \textbf{5. Determine} Pareto front of $S_i$ by passing $X_i$ and $Y_i$ to\\
    \hspace{20pt} Alg.~\ref{alg:mcdn_pareto}\\
    \hspace{20pt} \textbf{6. Remove} individuals in Pareto front of $S_i$ and add to $\hat{S}_i$\\
    \hspace{10pt} \textbf{7. Replace} $S_i$ with $\hat{S}_i$\\
    \hspace{10pt} \textbf{8. Truncate} $\hat{S}_i$ by removing the last fraction $F_l$\\
    \hspace{10pt} \textbf{8. Generate} new individuals using mutation/crossover\\
    \hspace{10pt} \textbf{9. Add} new individuals to $\hat{S}_i$, proceed as normal\\
\end{algorithm}

\begin{algorithm}[t]
\caption{Multiobj CoDeepNEAT Pareto Front Calculation}\label{alg:mcdn_pareto}
\flushleft
\textbf{1. Given} list of individuals $I$, and corresponding objective fitnesses $X$ and $Y$\\
\textbf{2. Sort} $I$ in descending order by first objective fitnesses $X$\\
\textbf{3. Create} new Pareto front PF with first individual $I_0$\\
\textbf{4. For each} individual $I_i$, $i>0$\\
    \hspace{10pt} \textbf{5. If} $Y_i$ is greater than the $Y_j$, where $I_j$ is last individual in PF\\
    \hspace{20pt} \textbf{6. Append} $I_i$ to PF\\
\textbf{7. Sort} PF in descending order by second objective $Y$ (Optional)\\
\end{algorithm}


\textit{DNN Complexity Minimization with Multiobjective Search.}
In addition to adapting CoDeepNEAT to multiple tasks, LEAF also extends it to multiple objectives.
In a single-objective evolutionary algorithm, elitism is applied to both the blueprint and the module populations.
The top fraction $F_l$ of the individuals within each species is passed on to the next generation as in single-objective optimization.
This fraction is based simply on ranking by fitness.
In the multiobjective version of CoDeepNEAT, the ranking is computed from successive Pareto fronts \cite{zhou2011multiobjective, deb2015multi} generated from the primary and secondary objectives.

Algorithm~\ref{alg:mcdn_lower} details this calculation for the blueprints and modules; assembled networks are also ranked similarly.
Algorithm~\ref{alg:mcdn_pareto} shows how the Pareto front, which is necessary for ranking, is calculated given a group of individuals that have been evaluated for each objective.
There is also an optional configuration parameter for multiobjective CoDeepNEAT that allows the individuals within each Pareto front to be sorted and ranked with respect to the secondary objective instead of the primary one.
Although the primary objective, i.e performance, is usually more important, this parameter can be used to emphasize the secondary objective more, if necessary for a particular domain.

Thus, multiobjective CoDeepNEAT can be used to maximize the performance and minimize the complexity of the evolved networks simultaneously.
While performance is usually measured as the loss on the unseen set of samples, there are many ways to characterize how complex a DNN is.
They include the number of parameters, the number of floating point operations (FLOPS), and the training/inference time of the network.
The most commonly used metric is number of parameters because other metrics can change depending on the deep learning library implementation and performance of the hardware.
In addition, this metric is becoming increasingly important in mobile applications as mobile devices are highly constrained in terms of memory and require networks with as high performance per parameter ratio as possible \cite{howard2017mobilenets}.
Thus, number of parameters is used as the secondary objective for multiobjective CoDeepNEAT in the experiments in the following section.
Although the number of parameters can vary widely across architectures, such variance does not pose a problem for multiobjective CoDeepNEAT since it only cares about the relative rankings between different objective values and no scaling of the secondary objective is required.

\section{Experimental Results}

LEAF's ability to democratize AI, improve the state-of-the-art, and minimize solutions is verified experimentally on two difficult real-world domains: (1) Wikipedia comment toxicity classification and (2) Chest X-rays multitask image classification.
The performance and efficiency of LEAF are compared against existing AutoML systems.

\subsection{Wikipedia Comment Toxicity Classification Domain}

Wikipedia is one of the largest encyclopedias that is publicly available online, with over 5 million written articles for the English language alone.
Unlike traditional encyclopedias, Wikipedia can be edited by any user who registers an account.
As a result, in the discussion section for some articles, there are often vitriolic or hateful comments that are directed at other users.
These comments are commonly referred to as ``toxic" and it has become increasingly important to detect toxic comments and remove them.
The Wikipedia Detox dataset (Wikidetox) is a collection of 160K example comments that are divided into 93K training, 31K validation, and 31K testing examples \cite{wikidetox}.
The labels for the comments are generated by humans using crowd-sourcing methods and contain four different categories for toxic comments.
However, following previous work \cite{chu2016comment}, all toxic comment categories are combined, thus creating a binary classification problem.
The dataset is also unbalanced with only about 9.6\% of the comments actually being labeled as toxic.

LEAF was configured to use standard CoDeepNEAT to search for well performing architectures in this domain.
The search space for these architectures was defined using recurrent (LSTM) layers as the basic building block.
Since comments are essentially an ordered list of words, recurrent layers (having been shown to be effective at processing sequential data \cite{mikolov2010recurrent}) were a natural choice.
In order for the words to be given as input into a recurrent network, they must be converted into an appropriate vector representation first.
Before given as input to the network, the comments were preprocessed using FastText, a recently introduced method for generating word embeddings \cite{bojanowski2016enriching} that improves upon the more commonly used Word2Vec \cite{mikolov2013distributed}.
Each evolved DNN was trained for three epochs and the classification accuracy on the testing set was returned as the fitness.
Preliminary experiments showed that three epochs of training was enough for the network performance to converge.
Thus, unlike vision domains such as Chest X-rays, there was no need for an extra step after evolution where the best evolved network was trained extensively from scratch.
The training and evaluation of networks at every generation were distributed over 100 worker machines.
For more information regarding evolution configuration and search space explored, refer to Tables~\ref{tb:experiment_params} and \ref{tb:search_space}, respectively.

\begin{figure*}[t]
  \begin{center}
    \includegraphics[width=0.95\linewidth]{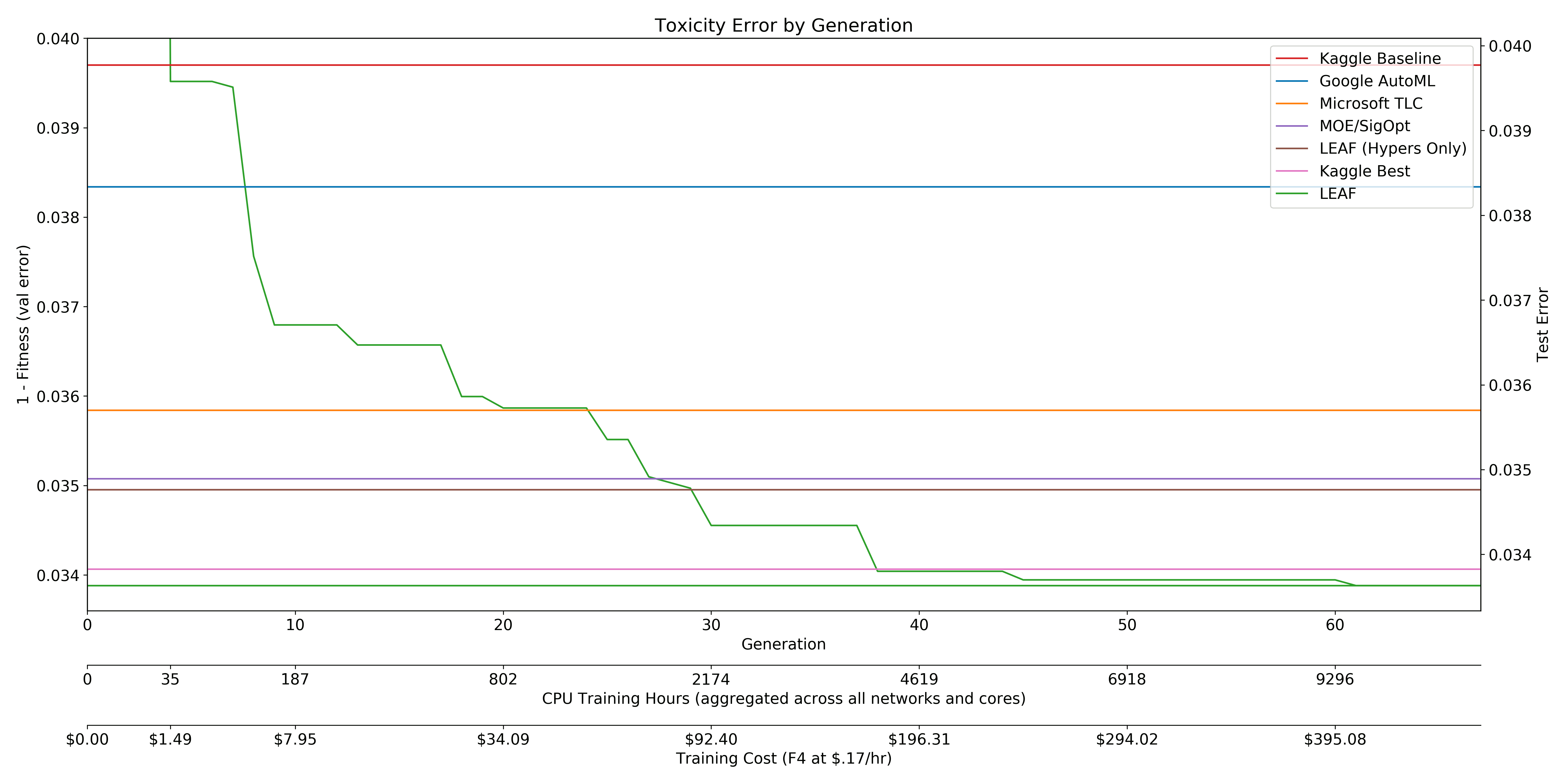}
    \caption{A comparison of LEAF in the Wikidetox domain against the networks discovered via several commercially available methods, including Keras, Google AutoML, MSFT TLC, and MOE, as well as the top human-designed network in the Kaggle comparison. The $y$-axis shows the best best fitness (i.e.\ accuracy) achieved so far, while the $x$-axes shows the generations, total training time, and total amount of money spent on cloud compute. As the plot shows, LEAF is gradually able to discover better networks, eventually finding one in the 40th generation that outperforms all other approaches.}
    \label{fig:wikidetox_results}
  \end{center}
\end{figure*}

The Wikidetox domain was part of a Kaggle challenge, and as a result, there already exists several hand-designed networks for the domain \cite{kaggle}.
Furthermore, due to the relative speed at which networks can be trained on this dataset, it was practical to evaluate hyperparameter optimization methods from companies such as Microsoft, Google, and MOE on this dataset.
Figure~\ref{fig:wikidetox_results} shows a comparison of LEAF architecture search against several other approaches.
The first one is the baseline network from the Kaggle competition, illustrating performance that a naive user can expect by applying a standard architecture to a new problem.
After spending just 35 CPU hours, LEAF found architectures that exceed that performance.
The next three comparisons illustrate the power of LEAF against other AutoML systems.
After about 180 hrs, it exceeded the performance of Google AutoML's text classifier optimization \cite{googleaiblog_2017} with default compute resources.
After about 1000 hrs, LEAF surpassed Microsoft's TLC library for model pipeline optimization \cite{tlc}, and after 2000 hrs, it exceeded MOE, a Bayesian optimization library \cite{moe} (both libraries used enough compute resources for the performance to plateau).
The LEAF hyperparameter-only version performed slightly better than MOE, demonstrating the power of evolution against other optimization approaches.
Finally, if the user is willing to spend about 9000 hrs of CPU time on this problem, the result is state-of-the-art performance.
At that point, LEAF discovered architectures that exceed the performance of Kaggle competition winner, i.e. improve upon the best known hand-design.
The performance gap between this result and the hyperparameter-only version of LEAF is also important: it shows the value added by optimizing network architectures, demonstrating that it is an essential ingredient in improving the state-of-the-art.

What is interesting about LEAF is that there are clear trade-offs between the amount of training time/money used and the quality of the results.
Depending on the budget available, the user running LEAF can stop earlier to get results competitive with existing approaches (such as TLC or Google AutomL) or run it to convergence to get the best possible results.
If the user is willing to spend more compute, increasingly more powerful architectures are obtained.
This kind of flexibility demonstrates that LEAF is not only a tool for improving AI, but also for democratizing AI.

\subsection{Chest X-rays Multitask Image Classification}

Chest X-rays classification is a recently introduced MTL benchmark \cite{wang2017chestx, rajpurkar2017chexnet}.
The dataset is composed of 112,120 high resolution frontal chest X-ray images, and the images are labeled with one or more of 14 different diseases, or no diseases at all.
The multi-label nature of the dataset naturally lends to an MTL setup where each disease is an individual binary classification task.
Past approaches generally apply the classical MTL DNN architecture \cite{wang2017chestx} and the current state-of-the-art approach uses a slightly modified version of Densenet \cite{rajpurkar2017chexnet}, a widely used, hand-designed architecture that is competitive with the state-of-the-art on the Imagenet domain \cite{huang2017densely}.
The images are divided into 70\% training, 10\% validation, and 20\% testing.
The metric used to evaluate the performance of the network is the average area under the ROC curve for all the tasks (AUROC).
Although the actual images are larger, all approaches (including LEAF) preprocessed the images to be $224\times224$ pixels, the same input size used by many Imagenet DNN architectures.

Since Chest X-rays is a multitask dataset, LEAF was configured to use the MTL variant of CoDeepNEAT to evolve network architectures.
The search space for these architectures was designed around 2D convolutional layers and includes skip connections seen in networks such as ResNet \cite{he2016deep}.
For fitness evaluations, all networks were trained using Adam \cite{Kingma:2014} for eight epochs.
After training was completed, AUROC was computed over all images in the validation set and returned as the fitness.
No data augmentation was performed during training and evaluation in evolution, but the images were normalized using the mean and variance statistics from the Imagenet dataset.
The average time for training was usually around 3-4 hours depending on the network size, although for some larger networks the training time exceeded 12 hours.
Like in the Wikidetox domain, the training and evaluation were parallelized over 100 worker machines.
For more information regarding evolution configuration and search space explored, refer to Tables~\ref{tb:experiment_params} and \ref{tb:search_space}, respectively.

After evolution converged, the best evolved network was trained for an increased number of epochs using the ADAM optimizer \cite{Kingma:2014}.
As with other approaches to neural architecture search \cite{zoph:arxiv16, real2018regularized}, the model augmentation method was used, where the number of filters of each convolutional layer was increased.
Data augmentation was also applied to the images during every epoch of training, including random horizontal flips, translations, and rotations.
The learning rate was dynamically adjusted downward based on the validation AUROC every epoch and sometimes reset back to its original value if the validation performance plateaued.
After training was complete, the testing set images were evaluated 20 times with data augmentation enabled and the network outputs were averaged to form the final prediction result.

\begin{table}[t]
\small
  \centering
  \begin{tabular}{lcc}
    \toprule
    \textbf{Algorithm} & \textbf{Test AUROC (\%)} \\
    \midrule
    1. Wang et al. (2017) \cite{wang2017chestx} & 73.8 \\
    2. CheXNet (2017) \cite{rajpurkar2017chexnet} & 84.4 \\
    3. Google AutoML (2018) \cite{googleaiblog_2017} & 79.7 \\
    \midrule
    4. LEAF & 84.3 \\
    \bottomrule
  \end{tabular}
  \caption{Performance on Chest X-rays testing set for hand-designed architectures and for networks that were evolved using Google AutoML and LEAF. LEAF improves significantly over Google AutoML and achieves performance virtually identically to the best hand-designed DNN, demonstrating state-of-the-art results in a task that requires very large networks.}
  \label{tb:chestxray_results}
\end{table}

\begin{figure*}[t]
  \begin{subfigure}[b]{0.32\linewidth}
    \centering
    \includegraphics[width=\linewidth]{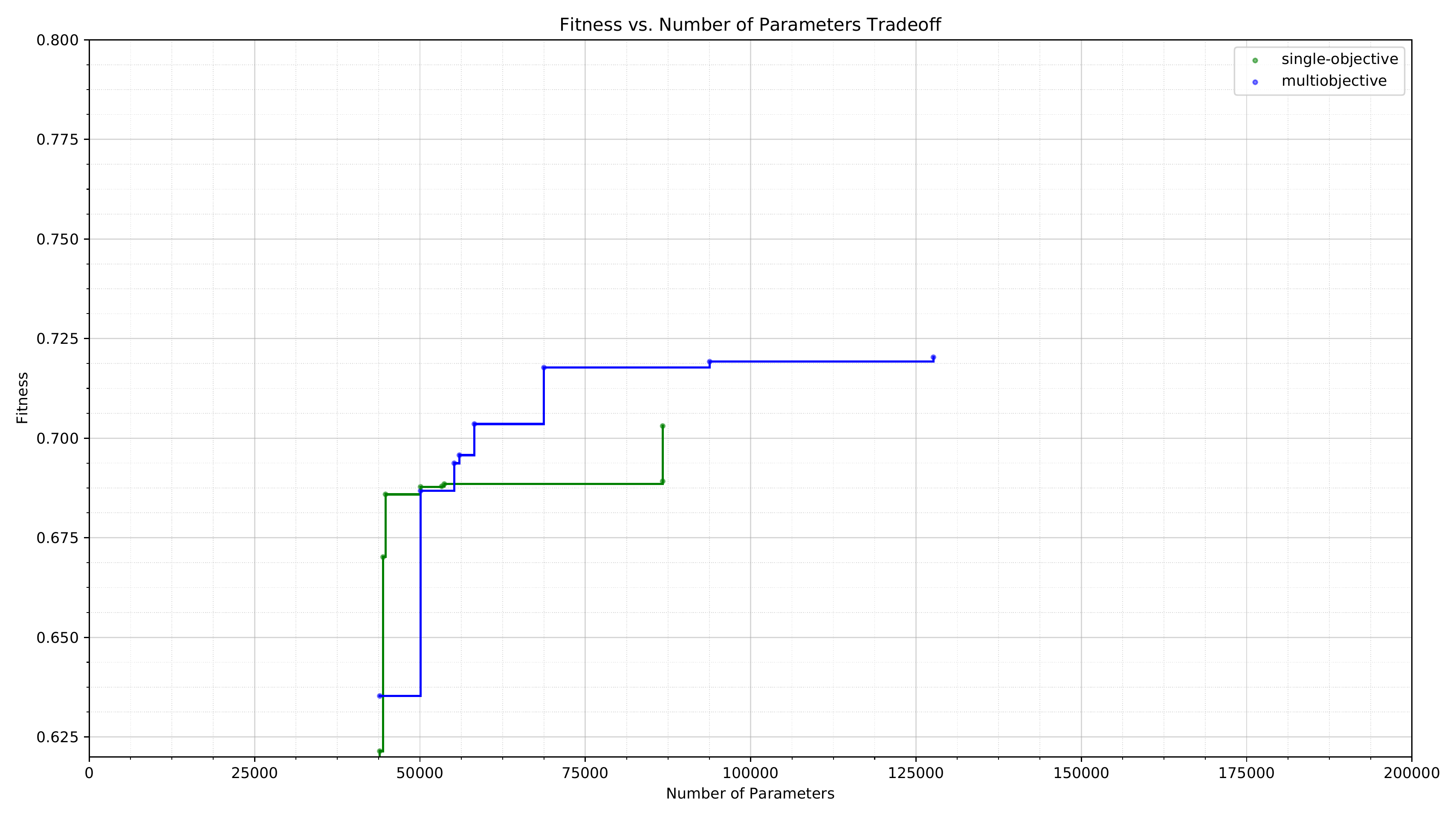}
    \caption{Generation 10}
  \end{subfigure}
  \begin{subfigure}[b]{0.32\linewidth}
    \centering
    \includegraphics[width=\linewidth]{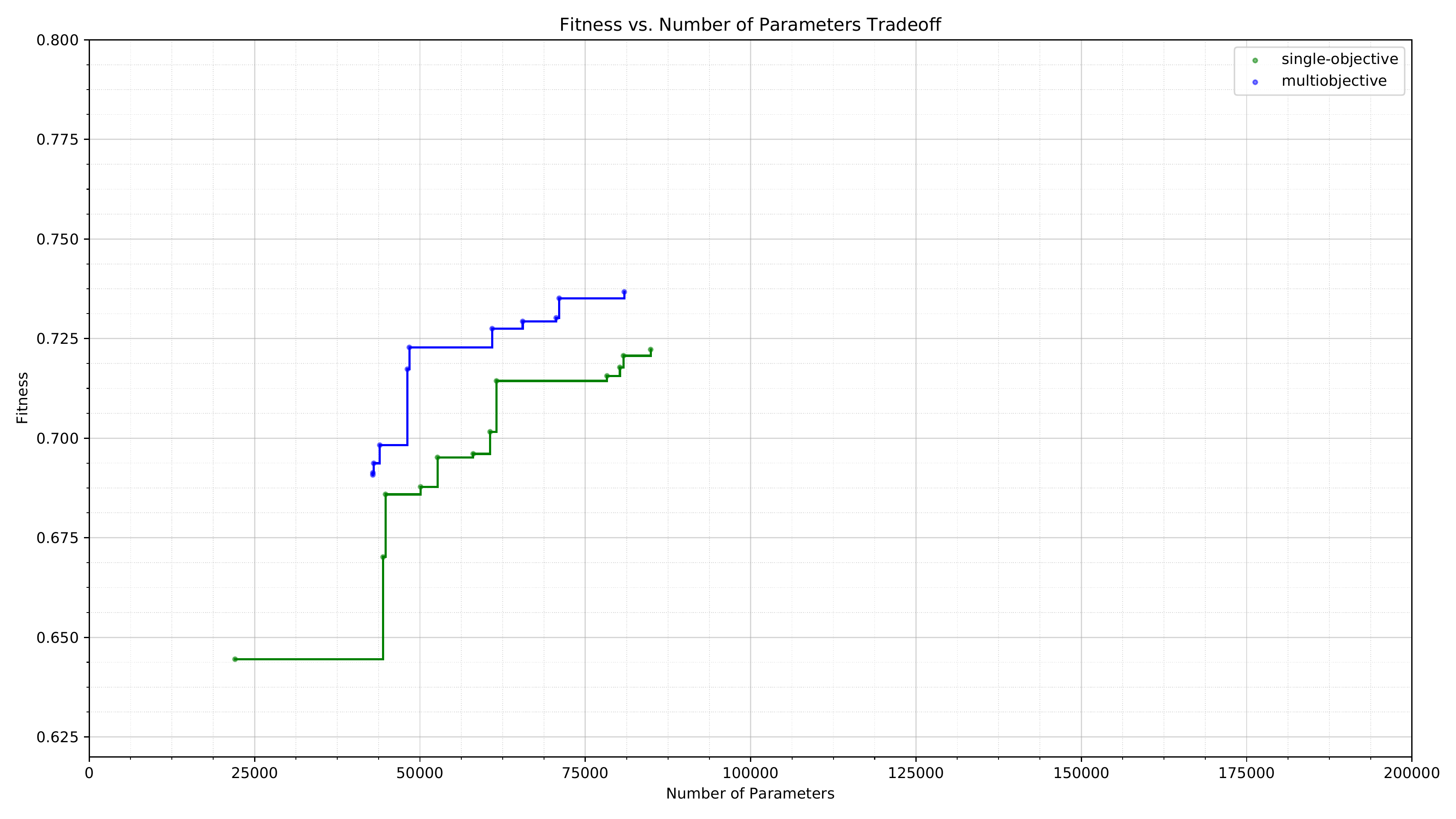}
    \caption{Generation 20}
  \end{subfigure}
  \begin{subfigure}[b]{0.32\linewidth}
    \centering
    \includegraphics[width=\linewidth]{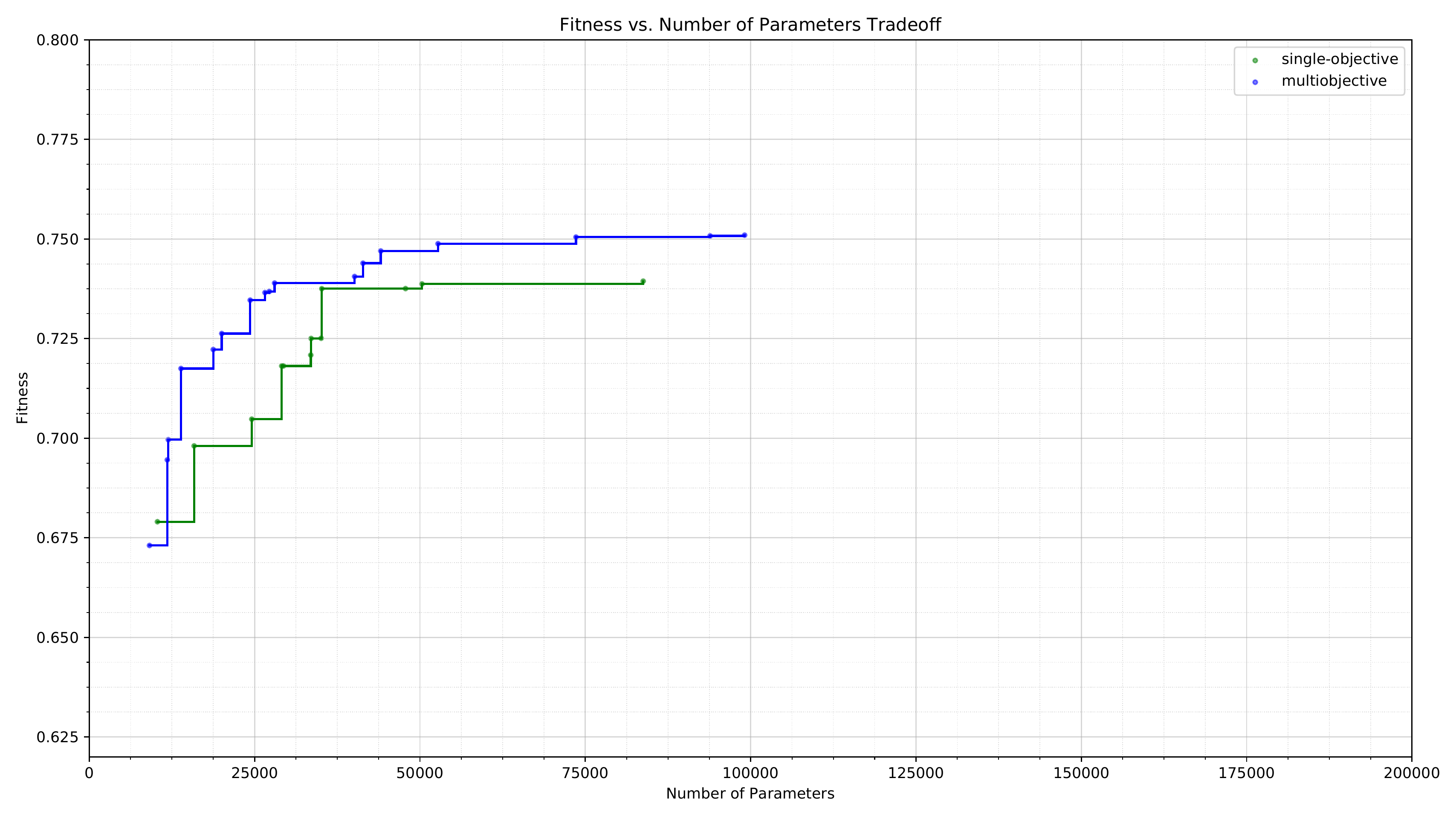}
    \caption{Generation 30}
  \end{subfigure}
  \begin{subfigure}[b]{0.32\linewidth}
    \centering
    \includegraphics[width=\linewidth]{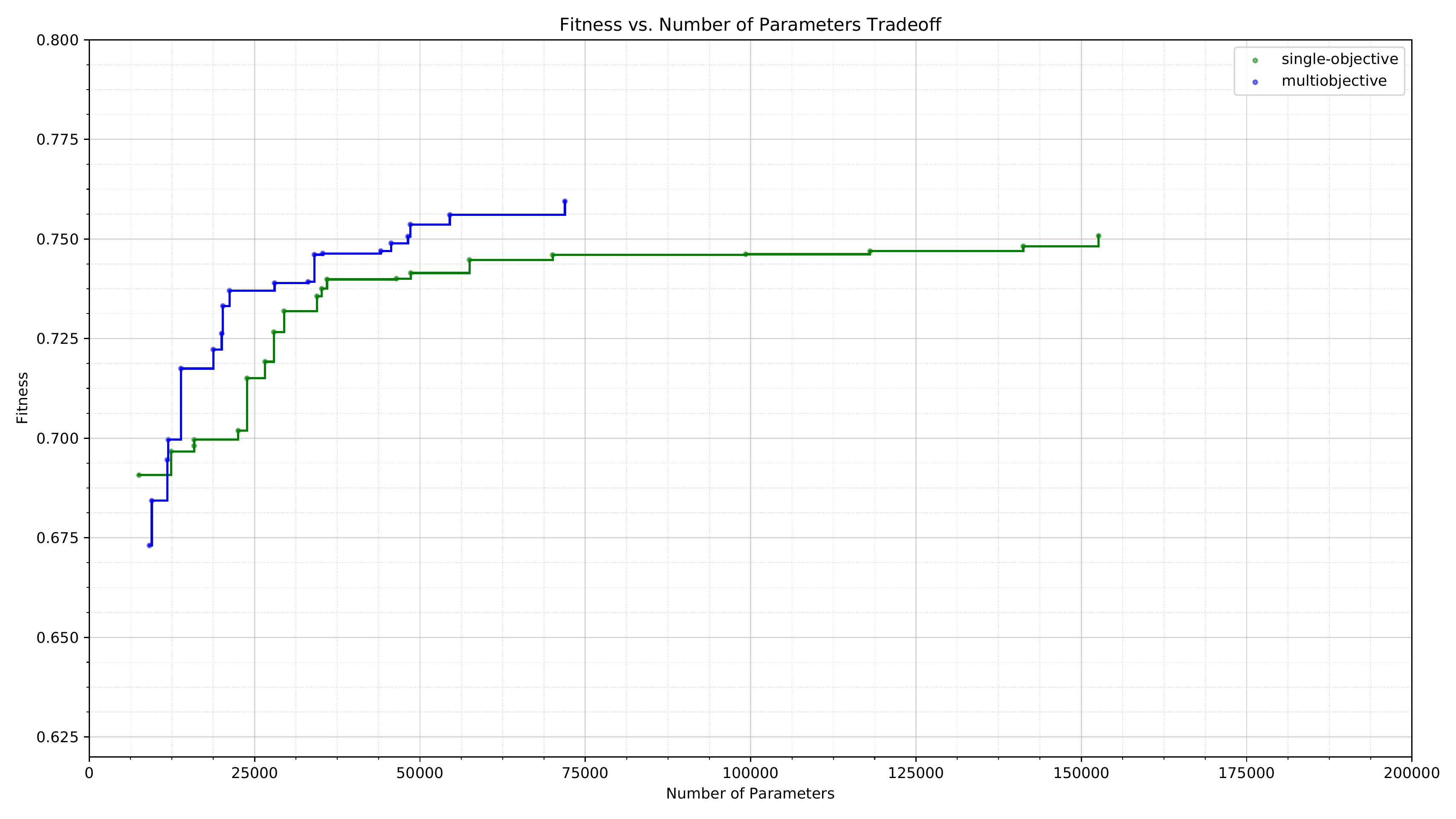}
    \caption{Generation 40}
  \end{subfigure}
  \begin{subfigure}[b]{0.32\linewidth}
    \centering
    \includegraphics[width=\linewidth]{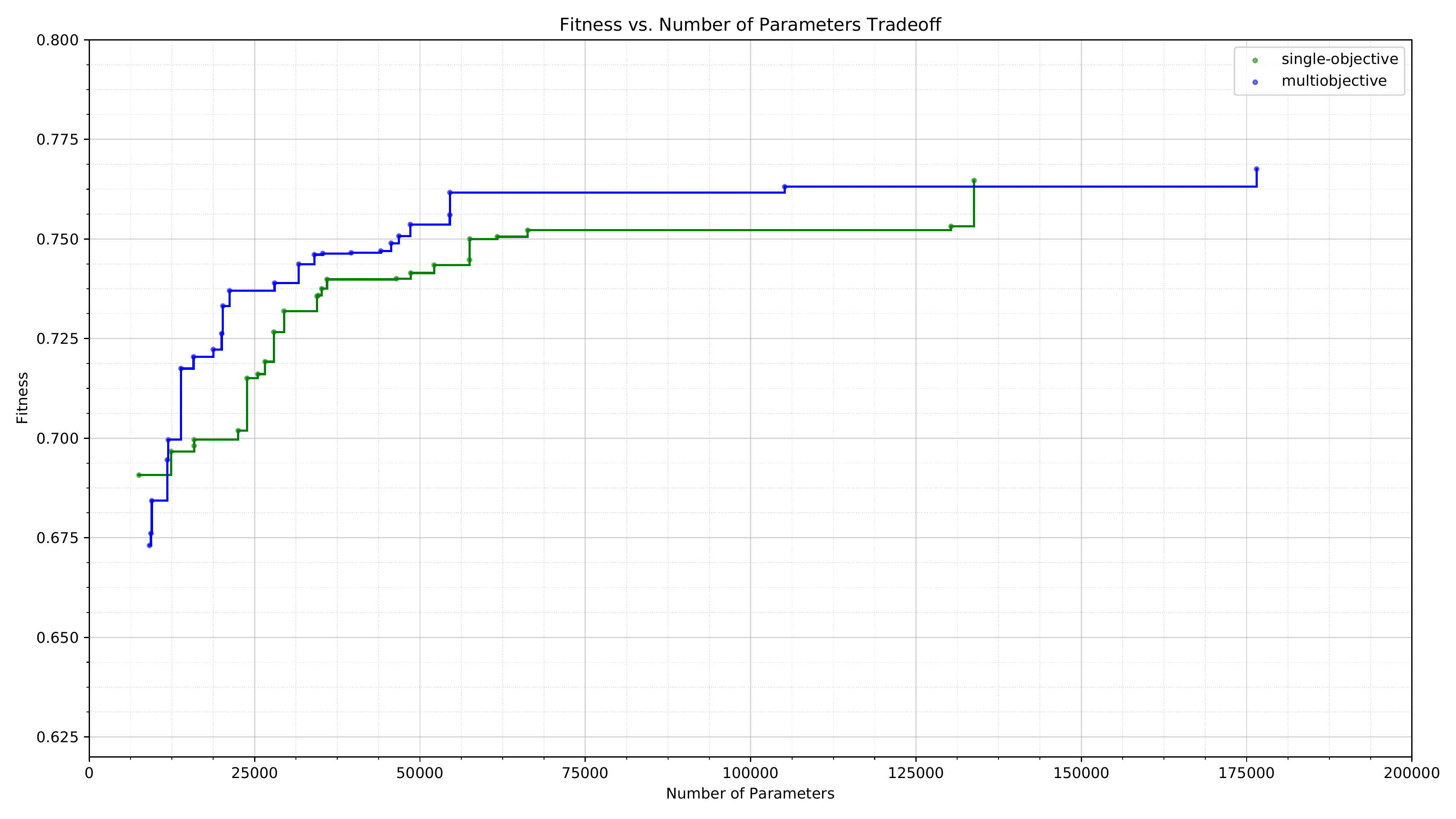}
    \caption{Generation 50}
  \end{subfigure}
  \begin{subfigure}[b]{0.32\linewidth}
    \centering
    \includegraphics[width=\linewidth]{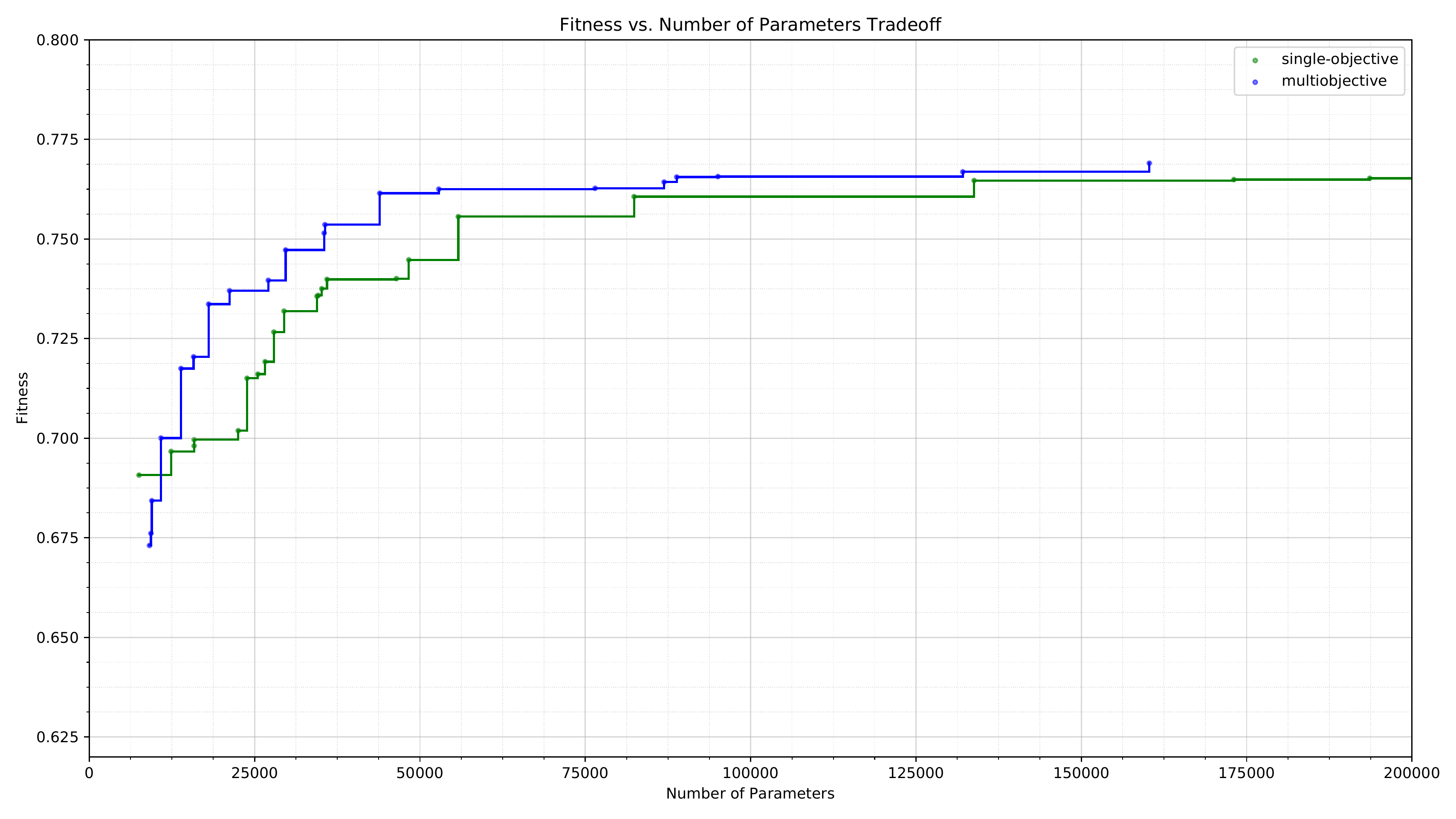}
    \caption{Generation 60}
  \end{subfigure}
  \caption{A comparison of Pareto fronts generated by LEAF using single-objective (green) and multiobjective (blue) CoDeepNEAT at various generations. The $x$-axis shows number of parameters (secondary objective) and the $y$-axis shows AUROC fitness (primary objective). The Pareto front for multiobjective LEAF dominates over the single objective Pareto front. In other words, multiobjective LEAF discovered trade-offs between complexity and performance that are always better than those found by standard, single-objective LEAF. Multiobjective LEAF not only found architectures with state-of-the-art performance, but also networks that are smaller and therefore more practical.}
  \label{fig:pareto_front}
\end{figure*}

Table~\ref{tb:chestxray_results} compares the performance of the best evolved networks with existing approaches that use hand-designed network architectures on a holdout testing set of images.
These include results from the authors who originally introduced the Chest X-rays dataset \cite{wang2017chestx} and also CheXNet \cite{rajpurkar2017chexnet}, which is the currently published state-of-the-art in this task.
For comparison with other AutoML systems, results from Google AutoML \cite{googleaiblog_2017} are also listed.
Google AutoML was set to optimize a vision task using a preset time of 24 hours (the higher of the two time limits available to the user), with an unknown amount of compute and number of worker machines.
Due to the size of the domain, it was not practical to evaluate Chest X-rays with other AutoML methods.
The performance of best network discovered by LEAF matches that of the human designed CheXNet.
LEAF is also able to exceed the performance of Google AutoML by a large margin of nearly 4 AUROC points.
These results demonstrate that state-of-the-art results are possible to achieve even in domains that require large, sophisticated networks.

An interesting question then is: can LEAF also minimize the size of these networks without sacrificing much in performance?
Interestingly, when LEAF used the multiobjective extension of CoDeepNEAT (multiobjective LEAF) to maximize fitness and minimize network size, LEAF actually converged faster during evolution to the same final fitness.
As expected, multiobjective LEAF was also able to discover networks with fewer parameters.
As shown in Figure~\ref{fig:pareto_front}, the Pareto front generated during evolution by multiobjective LEAF (blue) dominated that of single-objective LEAF (green) when compared at the same generations during evolution.
Although single-objective LEAF used standard CoDeepNEAT, it was possible to generate a Pareto front by giving the primary and secondary objective values of all the networks discovered in past generations to Algorithm~\ref{alg:mcdn_pareto}.
The Pareto front for multiobjective LEAF was also created the same way.

Interestingly, multiobjective LEAF discovered good networks in multiple phases.
In generation 10, networks found by these two approaches have similar average complexity but those evolved by multiobjective LEAF have much higher average fitness.
This situation changes later in evolution and by generation 40, the average complexity of the networks discovered by multiobjective LEAF is noticeably lower than that of single-objective LEAF, but the gap in average fitness between them has also narrowed.
Multiobjective LEAF first tried to optimize for the first objective (fitness) and only when fitness was starting to converge, did it try to improve the second objective (complexity).
In other words, multiobjective LEAF favored progress in the metric that was easiest to improve upon at the moment and did not get stuck; it would try to optimize another objective if no progress was made on the current one.

\begin{figure}[t]
  \begin{subfigure}[b]{0.35\linewidth}
    \centering
    \includegraphics[width=\linewidth]{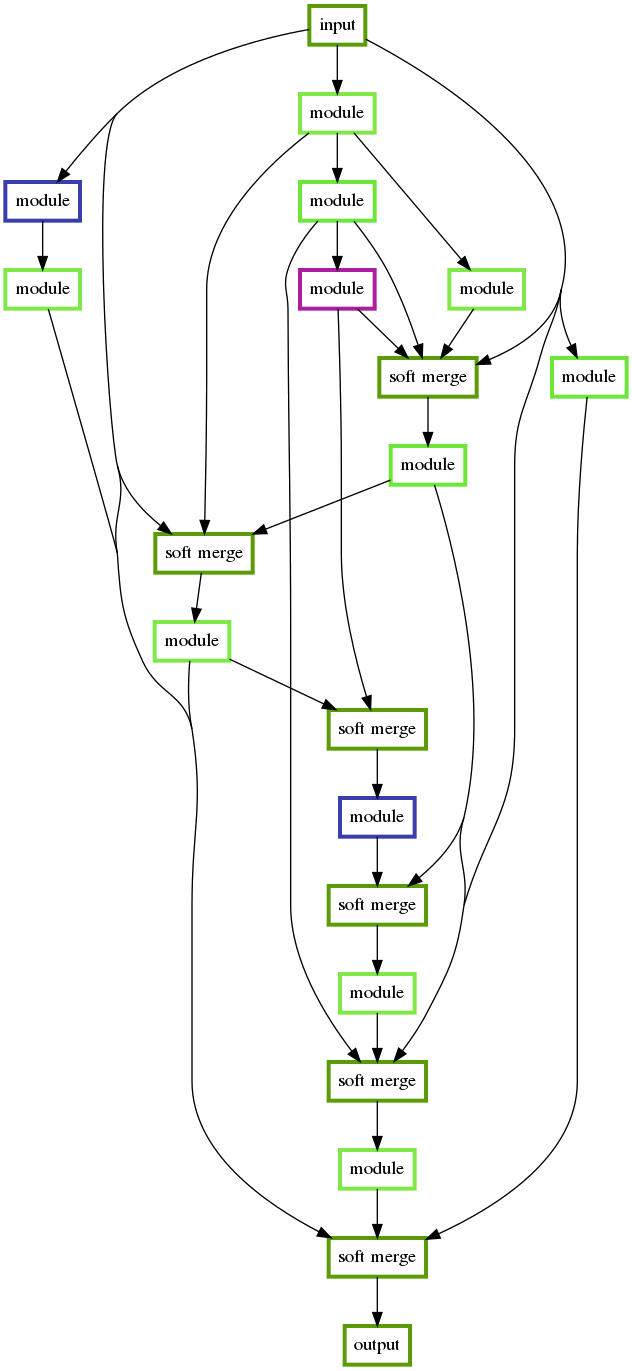}
    \caption{56K network}
    \label{fig:tradeoff_small}
  \end{subfigure}
  \begin{subfigure}[b]{0.35\linewidth}
    \centering
    \includegraphics[width=\linewidth]{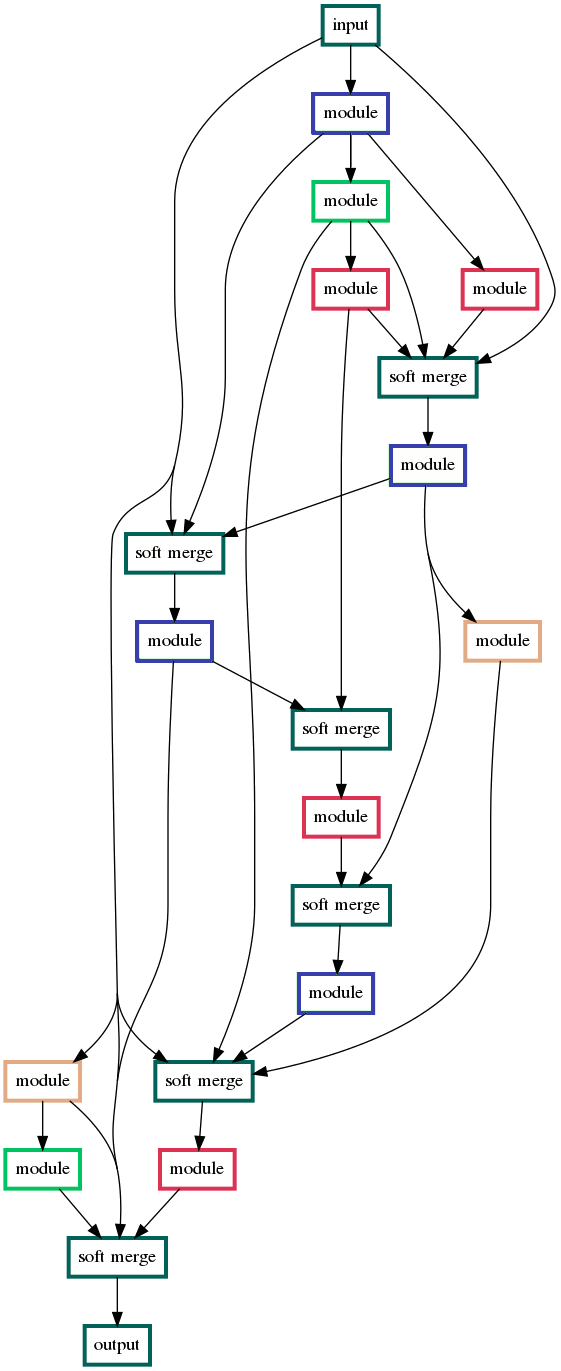}
    \caption{125K network}
    \label{fig:tradeoff_large}
  \end{subfigure}
  \begin{subfigure}[b]{0.2\linewidth}
    \centering
    \includegraphics[width=\linewidth]{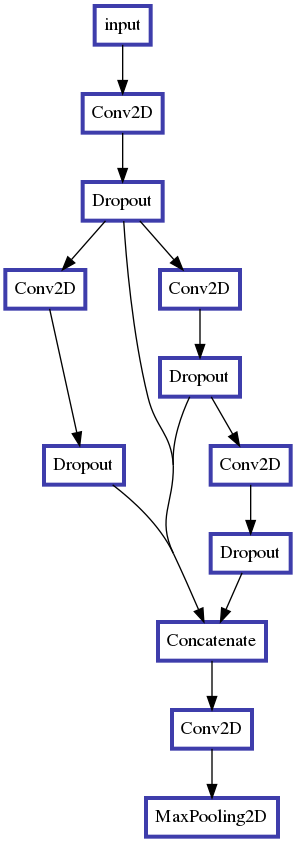}
    \caption{Module used by both networks}
    \label{fig:tradeoff_module}
  \end{subfigure}
  \caption{Visualizations of networks with different complexity discovered by multiobjective LEAF. The performance of the smaller 56K network (Figure~\ref{fig:tradeoff_small}) is nearly as good as that of the larger 125K network (Figure~\ref{fig:tradeoff_large}). The smaller network uses only two instances of the module architecture shown in Figure~\ref{fig:tradeoff_module} while the larger network uses four instances of the same module. These two networks show that multiobjective LEAF is able to find good trade-offs between two conflicting objectives by cleverly using modules.}
  \label{fig:mcdn_tradeoff}
\end{figure}

Visualizations of selected networks evolved by multiobjective LEAF are shown in Figure~\ref{fig:mcdn_tradeoff}.
LEAF was able to discover a very powerful network (Figure~\ref{fig:tradeoff_large}) that achieved 77\% AUROC after only eight epochs of training.
This network has 125K parameters and is already much smaller than networks with similar fitness discovered with single-objective LEAF.
Furthermore, multiobjective LEAF was able to discover an even smaller network (Figure~\ref{fig:tradeoff_small}) with only 56K parameters and a fitness of 74\% AUROC after eight epochs of training.
The main difference between the smaller and larger network is that the smaller one uses a particularly complex module architecture (Figure~\ref{fig:tradeoff_module}) only two times within its blueprint while the larger network uses the same module four times.
This result shows how a secondary objective can be used to bias the search towards smaller architectures without sacrificing much of the performance.

\section{Discussion}

The results for LEAF in the Wikidetox domain show that an evolutionary approach to optimizing deep neural networks is feasible.
It is possible to improve over a naive starting point with very little effort and to beat the existing state-of-the-art of both AutoML systems and hand-design with more effort.
Although a significant amount of compute is needed to train thousands of neural networks, it is promising that the results can be improved simply by running LEAF longer and by spending more on compute.
This feature is very useful in a commercial setting since it gives the user flexibility to find optimal trade-offs between resources spent and the quality of the results.
With computational power becoming cheaper and more available in the future, significantly better results are expected to be obtained.
Not all the approaches can put such power to good use, but evolutionary AutoML can.

The experiments with LEAF show that multiobjective optimization is effective in discovering networks that trade-off multiple metrics.
As seen in the Pareto fronts of Figure~\ref{fig:pareto_front}, the networks discovered by multiobjective LEAF dominate those evolved by single-objective LEAF with respect to the complexity and fitness metrics in almost every generation.
More surprisingly, multiobjective LEAF also maintains a higher average fitness with each generation.
This finding suggests that minimizing network complexity produces a regularization effect that also improves the generalization of the network.
This effect may be due to the fact that networks evolved by multiobjective LEAF reuse modules more often when compared to single-objective LEAF; extensive module reuse has been shown to improve performance in many hand-designed architectures \cite{szegedy2015going, he2016deep}.

In addition to the three goals of evolutionary AutoML demonstrated in this paper, a fourth one is to take advantage of multiple related datasets.
As shown in prior work \cite{Liang2018Evolutionary}, even when there is little data to train a DNN in a particular task, other tasks in a multitask setting can help achieve good performance.
Evolutionary AutoML thus forms a framework for utilizing DNNs in domains that otherwise would be impractical due to lack of data.

\section{Conclusion}

This paper showed that LEAF can outperform existing state-of-the-art AutoML systems and the best hand designed solutions.
The hyperparameters, components, and topology of the architecture can all be optimized simultaneously to fit the requirements of the task, resulting in superior performance.
LEAF achieves such results even if the user has little domain knowledge and provides a naive starting point, thus democratizing AI.
With LEAF, it is also possible to optimize other aspects of the architecture at the same time, such as size, making it more likely that the solutions discovered are useful in practice.

The biggest impact of automated design frameworks such as LEAF is that it makes new and unexpected applications of deep learning possible in vision, speech, language, robotics, and other areas.
In the long term, hand-design of algorithms and DNNs may be fully replaced by more sophisticated, general-purpose automated systems to aid scientists in their research or to aid engineers in designing AI-enabled products.

\section*{Appendix}

\begin{table}[h]
\footnotesize
  \centering
  \begin{tabular}{lcc}
    \toprule
    \textbf{Evolution parameter} & \textbf{Value} \\
    \midrule
    \textbf{Wikidetox domain} \\
    \midrule
    Module population size & 56 \\
    Add module node prob & 0.05 \\
    Add module connection prob & 0.05 \\
    Module species size & 4 \\
    Blueprint population size & 22 \\
    Add blueprint node prob & 0.05 \\
    Add blueprint connection prob & 0.05 \\
    Blueprint species size & 1 \\
    Assembled population size & 100 \\
    \midrule
    \textbf{Chest X-rays domain} \\
    \midrule
    Module population size & 56 \\
    Add module node prob & 0.08 \\
    Add module connection prob & 0.08 \\
    Module species size & 4 \\
    Blueprint population size & 22 \\
    Add blueprint node prob & 0.16 \\
    Add blueprint connection prob & 0.12 \\
    Blueprint species size & 1 \\
    Assembled population size & 100 \\
    \bottomrule
  \end{tabular}
  \caption{Evolution configuration for the Wikidetox and Chest X-rays domains.}
  \label{tb:experiment_params}
\end{table}

\begin{table}[h]
\footnotesize
  \centering
  \begin{tabular}{lcc}
    \toprule
    \textbf{Search space parameter} & \textbf{Range} \\
    \midrule
    \textbf{Wikidetox domain} \\
    \midrule
    Layer types & [Conv1D, LSTM, GRU, Dropout] \\
    Layer width & [64, 192] \\
    Kernel Size & [1, 3, 5, 7] \\
    Activation Function & [ReLU, Linear, ELU, SeLU] \\
    Weight Initializer & [Glorot, He] \\
    Dropout Rate & [0, 0.5] \\
    Weight Decay & [1e-9, 1e-3] \\
    Minimum pooling layers & 5 \\
    Weight sharing between layers & False \\
    \midrule
    \textbf{Chest X-rays domain} \\
    \midrule
    Layer types & [Conv2D, Dropout] \\
    Layer width & [16, 64] \\
    Kernel Size & [1, 3] \\
    Activation Function & [ReLU, Linear, ELU, SeLU] \\
    Weight Initializer & [Glorot, He] \\
    Dropout Rate & [0, 0.7] \\
    Weight Decay & [1e-9, 1e-3] \\
    Minimum pooling layers & 4 \\
    Weight sharing between layers & True \\
    \bottomrule
  \end{tabular}
  \caption{Search space explored for the Wikidetox and Chest X-rays domains.}
  \label{tb:search_space}
\end{table}

\section*{Acknowledgements}

Thanks to Joseph Sirosh and Justin Ormont for inspiring discussions and feedback in the course of this research, and in particular to Joseph for suggestions on minimization and hyperparameter optimization
comparisons, and to Justin for running the TLC comparisons.

\FloatBarrier

\bibliographystyle{ACM-Reference-Format}
\bibliography{mainbib,nn,risto}


\begin{thebibliography}{55}


\ifx \showCODEN    \undefined \def \showCODEN     #1{\unskip}     \fi
\ifx \showDOI      \undefined \def \showDOI       #1{#1}\fi
\ifx \showISBNx    \undefined \def \showISBNx     #1{\unskip}     \fi
\ifx \showISBNxiii \undefined \def \showISBNxiii  #1{\unskip}     \fi
\ifx \showISSN     \undefined \def \showISSN      #1{\unskip}     \fi
\ifx \showLCCN     \undefined \def \showLCCN      #1{\unskip}     \fi
\ifx \shownote     \undefined \def \shownote      #1{#1}          \fi
\ifx \showarticletitle \undefined \def \showarticletitle #1{#1}   \fi
\ifx \showURL      \undefined \def \showURL       {\relax}        \fi
\providecommand\bibfield[2]{#2}
\providecommand\bibinfo[2]{#2}
\providecommand\natexlab[1]{#1}
\providecommand\showeprint[2][]{arXiv:#2}

\bibitem[\protect\citeauthoryear{??}{goo}{2017}]%
        {googleaiblog_2017}
 \bibinfo{year}{2017}\natexlab{}.
\newblock \bibinfo{title}{AutoML for large scale image classification and
  object detection}.
\newblock
  \bibinfo{howpublished}{\url{https://ai.googleblog.com/2017/11/automl-for-large-scale-image.html}}.
    (\bibinfo{date}{Nov} \bibinfo{year}{2017}).
\newblock


\bibitem[\protect\citeauthoryear{??}{ama}{2019}]%
        {amazon}
 \bibinfo{year}{2019}\natexlab{}.
\newblock \bibinfo{title}{Amazon Web Services (AWS) - Cloud Computing
  Services}.
\newblock \bibinfo{howpublished}{\url{aws.amazon.com}}.
  (\bibinfo{year}{2019}).
\newblock


\bibitem[\protect\citeauthoryear{??}{gcl}{2019}]%
        {gcloud}
 \bibinfo{year}{2019}\natexlab{}.
\newblock \bibinfo{title}{Google Cloud}.
\newblock \bibinfo{howpublished}{\url{cloud.google.com}}.
  (\bibinfo{year}{2019}).
\newblock


\bibitem[\protect\citeauthoryear{??}{kag}{2019}]%
        {kaggle}
 \bibinfo{year}{2019}\natexlab{}.
\newblock \bibinfo{title}{Jigsaw Toxic Comment Classification Challenge}.
\newblock
  \bibinfo{howpublished}{\url{kaggle.com/jigsaw-toxic-comment-classification-challenge
  }}.   (\bibinfo{year}{2019}).
\newblock


\bibitem[\protect\citeauthoryear{??}{moe}{2019}]%
        {moe}
 \bibinfo{year}{2019}\natexlab{}.
\newblock \bibinfo{title}{Metric Optimization Engine}.
\newblock \bibinfo{howpublished}{\url{https://github.com/Yelp/MOE}}.
  (\bibinfo{year}{2019}).
\newblock


\bibitem[\protect\citeauthoryear{??}{azu}{2019}]%
        {azure}
 \bibinfo{year}{2019}\natexlab{}.
\newblock \bibinfo{title}{Microsoft Azure Cloud Computing Platform and
  Services}.
\newblock \bibinfo{howpublished}{\url{azure.microsoft.com}}.
  (\bibinfo{year}{2019}).
\newblock


\bibitem[\protect\citeauthoryear{??}{wik}{2019}]%
        {wikidetox}
 \bibinfo{year}{2019}\natexlab{}.
\newblock \bibinfo{title}{Research:Detox/Data Release}.
\newblock \bibinfo{howpublished}{\url{meta.wikimedia.org/wiki/Research:Detox}}.
    (\bibinfo{year}{2019}).
\newblock


\bibitem[\protect\citeauthoryear{??}{sig}{2019}]%
        {sigopt}
 \bibinfo{year}{2019}\natexlab{}.
\newblock \bibinfo{title}{SigOpt}.
\newblock \bibinfo{howpublished}{\url{sigopt.com/product}}.
  (\bibinfo{year}{2019}).
\newblock


\bibitem[\protect\citeauthoryear{??}{stu}{2019}]%
        {studioml}
 \bibinfo{year}{2019}\natexlab{}.
\newblock \bibinfo{title}{StudioML}.
\newblock \bibinfo{howpublished}{\url{https://www.studio.ml/}}.
  (\bibinfo{year}{2019}).
\newblock


\bibitem[\protect\citeauthoryear{??}{tlc}{2019}]%
        {tlc}
 \bibinfo{year}{2019}\natexlab{}.
\newblock \bibinfo{title}{Using the Microsoft TLC Machine Learning Tool}.
\newblock
  \bibinfo{howpublished}{\url{jamesmccaffrey.wordpress.com/2017/01/13/using-the-microsoft-tlc-machine-learning-tool
  }}.   (\bibinfo{year}{2019}).
\newblock


\bibitem[\protect\citeauthoryear{Bergstra and Bengio}{Bergstra and
  Bengio}{2012}]%
        {bergstra2012random}
\bibfield{author}{\bibinfo{person}{James Bergstra} {and}
  \bibinfo{person}{Yoshua Bengio}.} \bibinfo{year}{2012}\natexlab{}.
\newblock \showarticletitle{Random search for hyper-parameter optimization}.
\newblock \bibinfo{journal}{{\em Journal of Machine Learning Research\/}}
  \bibinfo{volume}{13}, \bibinfo{number}{Feb} (\bibinfo{year}{2012}),
  \bibinfo{pages}{281--305}.
\newblock


\bibitem[\protect\citeauthoryear{Bojanowski, Grave, Joulin, and
  Mikolov}{Bojanowski et~al\mbox{.}}{2016}]%
        {bojanowski2016enriching}
\bibfield{author}{\bibinfo{person}{Piotr Bojanowski}, \bibinfo{person}{Edouard
  Grave}, \bibinfo{person}{Armand Joulin}, {and} \bibinfo{person}{Tomas
  Mikolov}.} \bibinfo{year}{2016}\natexlab{}.
\newblock \showarticletitle{Enriching Word Vectors with Subword Information}.
\newblock \bibinfo{journal}{{\em arXiv preprint arXiv:1607.04606\/}}
  (\bibinfo{year}{2016}).
\newblock


\bibitem[\protect\citeauthoryear{Che, Purushotham, Cho, Sontag, and Liu}{Che
  et~al\mbox{.}}{2016}]%
        {che:arxiv16}
\bibfield{author}{\bibinfo{person}{Zhengping Che}, \bibinfo{person}{Sanjay
  Purushotham}, \bibinfo{person}{Kyunghyun Cho}, \bibinfo{person}{David
  Sontag}, {and} \bibinfo{person}{Yan Liu}.} \bibinfo{year}{2016}\natexlab{}.
\newblock \showarticletitle{Recurrent Neural Networks for Multivariate Time
  Series with Missing Values}.
\newblock \bibinfo{journal}{{\em CoRR\/}}  \bibinfo{volume}{abs/1606.01865}
  (\bibinfo{year}{2016}).
\newblock
\showURL{%
\url{http://arxiv.org/abs/1606.01865}}


\bibitem[\protect\citeauthoryear{Chollet et~al\mbox{.}}{Chollet
  et~al\mbox{.}}{2015}]%
        {Chollet:2015}
\bibfield{author}{\bibinfo{person}{F. Chollet} {et~al\mbox{.}}}
  \bibinfo{year}{2015}\natexlab{}.
\newblock \bibinfo{title}{Keras}.
\newblock   (\bibinfo{year}{2015}).
\newblock


\bibitem[\protect\citeauthoryear{Chu, Jue, and Wang}{Chu et~al\mbox{.}}{2016}]%
        {chu2016comment}
\bibfield{author}{\bibinfo{person}{Theodora Chu}, \bibinfo{person}{Kylie Jue},
  {and} \bibinfo{person}{Max Wang}.} \bibinfo{year}{2016}\natexlab{}.
\newblock \showarticletitle{Comment Abuse Classification with Deep Learning}.
\newblock \bibinfo{journal}{{\em Von https://web. stanford.
  edu/class/cs224n/reports/2762092. pdf abgerufen\/}} (\bibinfo{year}{2016}).
\newblock


\bibitem[\protect\citeauthoryear{Collobert and Weston}{Collobert and
  Weston}{2008}]%
        {collobert:icml08}
\bibfield{author}{\bibinfo{person}{Ronan Collobert} {and}
  \bibinfo{person}{Jason Weston}.} \bibinfo{year}{2008}\natexlab{}.
\newblock \showarticletitle{A unified architecture for natural language
  processing: {D}eep neural networks with multitask learning}. In
  \bibinfo{booktitle}{{\em Proceedings of the 25th international conference on
  Machine learning}}. \bibinfo{publisher}{ACM}, \bibinfo{pages}{160--167}.
\newblock


\bibitem[\protect\citeauthoryear{Deb}{Deb}{2015}]%
        {deb2015multi}
\bibfield{author}{\bibinfo{person}{Kalyanmoy Deb}.}
  \bibinfo{year}{2015}\natexlab{}.
\newblock \showarticletitle{Multi-objective evolutionary algorithms}.
\newblock In \bibinfo{booktitle}{{\em Springer Handbook of Computational
  Intelligence}}. \bibinfo{publisher}{Springer}, \bibinfo{pages}{995--1015}.
\newblock


\bibitem[\protect\citeauthoryear{Elsken, Metzen, and Hutter}{Elsken
  et~al\mbox{.}}{1804}]%
        {elsken1804efficient}
\bibfield{author}{\bibinfo{person}{Thomas Elsken}, \bibinfo{person}{J~Hendrik
  Metzen}, {and} \bibinfo{person}{Frank Hutter}.}
  \bibinfo{year}{1804}\natexlab{}.
\newblock \showarticletitle{Efficient Multi-objective Neural Architecture
  Search via Lamarckian Evolution}.
\newblock \bibinfo{journal}{{\em ArXiv e-prints\/}} (\bibinfo{year}{1804}).
\newblock


\bibitem[\protect\citeauthoryear{Gomez, Schmidhuber, and Miikkulainen}{Gomez
  et~al\mbox{.}}{2008}]%
        {gomez2008accelerated}
\bibfield{author}{\bibinfo{person}{Faustino Gomez}, \bibinfo{person}{J{\"u}rgen
  Schmidhuber}, {and} \bibinfo{person}{Risto Miikkulainen}.}
  \bibinfo{year}{2008}\natexlab{}.
\newblock \showarticletitle{Accelerated neural evolution through cooperatively
  coevolved synapses}.
\newblock \bibinfo{journal}{{\em Journal of Machine Learning Research\/}}
  \bibinfo{volume}{9}, \bibinfo{number}{May} (\bibinfo{year}{2008}),
  \bibinfo{pages}{937--965}.
\newblock


\bibitem[\protect\citeauthoryear{Gomez and Miikkulainen}{Gomez and
  Miikkulainen}{1999}]%
        {gomez1999solving}
\bibfield{author}{\bibinfo{person}{Faustino~J Gomez} {and}
  \bibinfo{person}{Risto Miikkulainen}.} \bibinfo{year}{1999}\natexlab{}.
\newblock \showarticletitle{Solving non-Markovian control tasks with
  neuroevolution}. In \bibinfo{booktitle}{{\em IJCAI}},
  Vol.~\bibinfo{volume}{99}. \bibinfo{pages}{1356--1361}.
\newblock


\bibitem[\protect\citeauthoryear{Graves, Mohamed, and Hinton}{Graves
  et~al\mbox{.}}{2013}]%
        {graves:icassp13}
\bibfield{author}{\bibinfo{person}{Alex Graves}, \bibinfo{person}{Abdel-rahman
  Mohamed}, {and} \bibinfo{person}{Geoffrey Hinton}.}
  \bibinfo{year}{2013}\natexlab{}.
\newblock \showarticletitle{Speech recognition with deep recurrent neural
  networks}. In \bibinfo{booktitle}{{\em 2013 {IEEE} International Conference
  on Acoustics, Speech and Signal Processing}}. \bibinfo{publisher}{IEEE},
  \bibinfo{pages}{6645--6649}.
\newblock


\bibitem[\protect\citeauthoryear{He, Zhang, Ren, and Sun}{He
  et~al\mbox{.}}{2016a}]%
        {he2016deep}
\bibfield{author}{\bibinfo{person}{Kaiming He}, \bibinfo{person}{Xiangyu
  Zhang}, \bibinfo{person}{Shaoqing Ren}, {and} \bibinfo{person}{Jian Sun}.}
  \bibinfo{year}{2016}\natexlab{a}.
\newblock \showarticletitle{Deep residual learning for image recognition}. In
  \bibinfo{booktitle}{{\em Proceedings of the IEEE Conference on Computer
  Vision and Pattern Recognition}}. \bibinfo{pages}{770--778}.
\newblock


\bibitem[\protect\citeauthoryear{He, Zhang, Ren, and Sun}{He
  et~al\mbox{.}}{2016b}]%
        {he:arxiv16}
\bibfield{author}{\bibinfo{person}{Kaiming He}, \bibinfo{person}{Xiangyu
  Zhang}, \bibinfo{person}{Shaoqing Ren}, {and} \bibinfo{person}{Jian Sun}.}
  \bibinfo{year}{2016}\natexlab{b}.
\newblock \showarticletitle{Identity Mappings in Deep Residual Networks}.
\newblock \bibinfo{journal}{{\em CoRR\/}}  \bibinfo{volume}{abs/1603.05027}
  (\bibinfo{year}{2016}).
\newblock
\showURL{%
\url{http://arxiv.org/abs/1603.05027}}


\bibitem[\protect\citeauthoryear{He, Zhang, Ren, and Sun}{He
  et~al\mbox{.}}{2016c}]%
        {he2016identity}
\bibfield{author}{\bibinfo{person}{Kaiming He}, \bibinfo{person}{Xiangyu
  Zhang}, \bibinfo{person}{Shaoqing Ren}, {and} \bibinfo{person}{Jian Sun}.}
  \bibinfo{year}{2016}\natexlab{c}.
\newblock \showarticletitle{Identity mappings in deep residual networks}. In
  \bibinfo{booktitle}{{\em European conference on computer vision}}. Springer,
  \bibinfo{pages}{630--645}.
\newblock


\bibitem[\protect\citeauthoryear{Howard, Zhu, Chen, Kalenichenko, Wang, Weyand,
  Andreetto, and Adam}{Howard et~al\mbox{.}}{2017}]%
        {howard2017mobilenets}
\bibfield{author}{\bibinfo{person}{Andrew~G Howard}, \bibinfo{person}{Menglong
  Zhu}, \bibinfo{person}{Bo Chen}, \bibinfo{person}{Dmitry Kalenichenko},
  \bibinfo{person}{Weijun Wang}, \bibinfo{person}{Tobias Weyand},
  \bibinfo{person}{Marco Andreetto}, {and} \bibinfo{person}{Hartwig Adam}.}
  \bibinfo{year}{2017}\natexlab{}.
\newblock \showarticletitle{Mobilenets: Efficient convolutional neural networks
  for mobile vision applications}.
\newblock \bibinfo{journal}{{\em arXiv preprint arXiv:1704.04861\/}}
  (\bibinfo{year}{2017}).
\newblock


\bibitem[\protect\citeauthoryear{Huang, Liu, Van Der~Maaten, and
  Weinberger}{Huang et~al\mbox{.}}{2017}]%
        {huang2017densely}
\bibfield{author}{\bibinfo{person}{Gao Huang}, \bibinfo{person}{Zhuang Liu},
  \bibinfo{person}{Laurens Van Der~Maaten}, {and} \bibinfo{person}{Kilian~Q
  Weinberger}.} \bibinfo{year}{2017}\natexlab{}.
\newblock \showarticletitle{Densely Connected Convolutional Networks.}. In
  \bibinfo{booktitle}{{\em CVPR}}, Vol.~\bibinfo{volume}{1}.
  \bibinfo{pages}{3}.
\newblock


\bibitem[\protect\citeauthoryear{Keogh and Mueen}{Keogh and Mueen}{2011}]%
        {keogh2011curse}
\bibfield{author}{\bibinfo{person}{Eamonn Keogh} {and}
  \bibinfo{person}{Abdullah Mueen}.} \bibinfo{year}{2011}\natexlab{}.
\newblock \showarticletitle{Curse of dimensionality}.
\newblock In \bibinfo{booktitle}{{\em Encyclopedia of machine learning}}.
  \bibinfo{publisher}{Springer}, \bibinfo{pages}{257--258}.
\newblock


\bibitem[\protect\citeauthoryear{Kingma and Ba}{Kingma and Ba}{2014}]%
        {Kingma:2014}
\bibfield{author}{\bibinfo{person}{D.~P. Kingma} {and} \bibinfo{person}{J.
  Ba}.} \bibinfo{year}{2014}\natexlab{}.
\newblock \showarticletitle{Adam: {A} Method for Stochastic Optimization}.
\newblock \bibinfo{journal}{{\em CoRR\/}}  \bibinfo{volume}{abs/1412.6980}
  (\bibinfo{year}{2014}).
\newblock


\bibitem[\protect\citeauthoryear{LeCun, Bengio, and Hinton}{LeCun
  et~al\mbox{.}}{2015}]%
        {lecun2015deep}
\bibfield{author}{\bibinfo{person}{Yann LeCun}, \bibinfo{person}{Yoshua
  Bengio}, {and} \bibinfo{person}{Geoffrey Hinton}.}
  \bibinfo{year}{2015}\natexlab{}.
\newblock \showarticletitle{Deep learning}.
\newblock \bibinfo{journal}{{\em Nature\/}} \bibinfo{volume}{521},
  \bibinfo{number}{7553} (\bibinfo{year}{2015}), \bibinfo{pages}{436--444}.
\newblock


\bibitem[\protect\citeauthoryear{Liang, Meyerson, and Miikkulainen}{Liang
  et~al\mbox{.}}{2018}]%
        {Liang2018Evolutionary}
\bibfield{author}{\bibinfo{person}{Jason Liang}, \bibinfo{person}{Elliot
  Meyerson}, {and} \bibinfo{person}{Risto Miikkulainen}.}
  \bibinfo{year}{2018}\natexlab{}.
\newblock \showarticletitle{Evolutionary Architecture Search for Deep Multitask
  Networks}. In \bibinfo{booktitle}{{\em Proceedings of the Genetic and
  Evolutionary Computation Conference}} {\em (\bibinfo{series}{GECCO '18})}.
  \bibinfo{publisher}{ACM}, \bibinfo{address}{New York, NY, USA},
  \bibinfo{pages}{466--473}.
\newblock
\showISBNx{978-1-4503-5618-3}
\showDOI{%
\url{https://doi.org/10.1145/3205455.3205489}}


\bibitem[\protect\citeauthoryear{Liu, Simonyan, Vinyals, Fernando, and
  Kavukcuoglu}{Liu et~al\mbox{.}}{2017}]%
        {liu2017hierarchical}
\bibfield{author}{\bibinfo{person}{Hanxiao Liu}, \bibinfo{person}{Karen
  Simonyan}, \bibinfo{person}{Oriol Vinyals}, \bibinfo{person}{Chrisantha
  Fernando}, {and} \bibinfo{person}{Koray Kavukcuoglu}.}
  \bibinfo{year}{2017}\natexlab{}.
\newblock \showarticletitle{Hierarchical representations for efficient
  architecture search}.
\newblock \bibinfo{journal}{{\em arXiv preprint arXiv:1711.00436\/}}
  (\bibinfo{year}{2017}).
\newblock


\bibitem[\protect\citeauthoryear{Loshchilov and Hutter}{Loshchilov and
  Hutter}{2016}]%
        {loshchilov2016cma}
\bibfield{author}{\bibinfo{person}{Ilya Loshchilov} {and}
  \bibinfo{person}{Frank Hutter}.} \bibinfo{year}{2016}\natexlab{}.
\newblock \showarticletitle{CMA-ES for Hyperparameter Optimization of Deep
  Neural Networks}.
\newblock \bibinfo{journal}{{\em arXiv preprint arXiv:1604.07269\/}}
  (\bibinfo{year}{2016}).
\newblock


\bibitem[\protect\citeauthoryear{Lu, Whalen, Boddeti, Dhebar, Deb, Goodman, and
  Banzhaf}{Lu et~al\mbox{.}}{2018}]%
        {lu2018nsga}
\bibfield{author}{\bibinfo{person}{Zhichao Lu}, \bibinfo{person}{Ian Whalen},
  \bibinfo{person}{Vishnu Boddeti}, \bibinfo{person}{Yashesh Dhebar},
  \bibinfo{person}{Kalyanmoy Deb}, \bibinfo{person}{Erik Goodman}, {and}
  \bibinfo{person}{Wolfgang Banzhaf}.} \bibinfo{year}{2018}\natexlab{}.
\newblock \showarticletitle{NSGA-NET: A Multi-Objective Genetic Algorithm for
  Neural Architecture Search}.
\newblock \bibinfo{journal}{{\em arXiv preprint arXiv:1810.03522\/}}
  (\bibinfo{year}{2018}).
\newblock


\bibitem[\protect\citeauthoryear{Meyerson and Miikkulainen}{Meyerson and
  Miikkulainen}{2018}]%
        {Meyerson:2018}
\bibfield{author}{\bibinfo{person}{E. Meyerson} {and} \bibinfo{person}{R.
  Miikkulainen}.} \bibinfo{year}{2018}\natexlab{}.
\newblock \showarticletitle{Beyond Shared Hierarchies: Deep Multitask Learning
  through Soft Layer Ordering}.
\newblock \bibinfo{journal}{{\em ICLR\/}} (\bibinfo{year}{2018}).
\newblock


\bibitem[\protect\citeauthoryear{Miikkulainen, Liang, Meyerson,
  et~al\mbox{.}}{Miikkulainen et~al\mbox{.}}{2017}]%
        {miikkulainen2017evolving}
\bibfield{author}{\bibinfo{person}{R. Miikkulainen}, \bibinfo{person}{J.
  Liang}, \bibinfo{person}{E. Meyerson}, {et~al\mbox{.}}}
  \bibinfo{year}{2017}\natexlab{}.
\newblock \showarticletitle{Evolving deep neural networks}.
\newblock \bibinfo{journal}{{\em arXiv preprint arXiv:1703.00548\/}}
  (\bibinfo{year}{2017}).
\newblock


\bibitem[\protect\citeauthoryear{Mikolov, Karafi{\'a}t, Burget,
  {\v{C}}ernock{\`y}, and Khudanpur}{Mikolov et~al\mbox{.}}{2010}]%
        {mikolov2010recurrent}
\bibfield{author}{\bibinfo{person}{Tom{\'a}{\v{s}} Mikolov},
  \bibinfo{person}{Martin Karafi{\'a}t}, \bibinfo{person}{Luk{\'a}{\v{s}}
  Burget}, \bibinfo{person}{Jan {\v{C}}ernock{\`y}}, {and}
  \bibinfo{person}{Sanjeev Khudanpur}.} \bibinfo{year}{2010}\natexlab{}.
\newblock \showarticletitle{Recurrent neural network based language model}. In
  \bibinfo{booktitle}{{\em Eleventh Annual Conference of the International
  Speech Communication Association}}.
\newblock


\bibitem[\protect\citeauthoryear{Mikolov, Sutskever, Chen, Corrado, and
  Dean}{Mikolov et~al\mbox{.}}{2013}]%
        {mikolov2013distributed}
\bibfield{author}{\bibinfo{person}{Tomas Mikolov}, \bibinfo{person}{Ilya
  Sutskever}, \bibinfo{person}{Kai Chen}, \bibinfo{person}{Greg~S Corrado},
  {and} \bibinfo{person}{Jeff Dean}.} \bibinfo{year}{2013}\natexlab{}.
\newblock \showarticletitle{Distributed representations of words and phrases
  and their compositionality}. In \bibinfo{booktitle}{{\em Advances in neural
  information processing systems}}. \bibinfo{pages}{3111--3119}.
\newblock


\bibitem[\protect\citeauthoryear{Moriarty and Miikkulainen}{Moriarty and
  Miikkulainen}{1998}]%
        {moriarty1998hierarchical}
\bibfield{author}{\bibinfo{person}{David~E Moriarty} {and}
  \bibinfo{person}{Risto Miikkulainen}.} \bibinfo{year}{1998}\natexlab{}.
\newblock \showarticletitle{Hierarchical evolution of neural networks}. In
  \bibinfo{booktitle}{{\em Evolutionary Computation Proceedings, 1998. IEEE
  World Congress on Computational Intelligence., The 1998 IEEE International
  Conference on}}. IEEE, \bibinfo{pages}{428--433}.
\newblock


\bibitem[\protect\citeauthoryear{Ng, Hausknecht, Vijayanarasimhan, Vinyals,
  Monga, and Toderici}{Ng et~al\mbox{.}}{2015}]%
        {ng:arxiv15}
\bibfield{author}{\bibinfo{person}{Joe~Yue{-}Hei Ng},
  \bibinfo{person}{Matthew~J. Hausknecht}, \bibinfo{person}{Sudheendra
  Vijayanarasimhan}, \bibinfo{person}{Oriol Vinyals}, \bibinfo{person}{Rajat
  Monga}, {and} \bibinfo{person}{George Toderici}.}
  \bibinfo{year}{2015}\natexlab{}.
\newblock \showarticletitle{Beyond Short Snippets: Deep Networks for Video
  Classification}.
\newblock \bibinfo{journal}{{\em CoRR\/}}  \bibinfo{volume}{abs/1503.08909}
  (\bibinfo{year}{2015}).
\newblock
\showURL{%
\url{http://arxiv.org/abs/1503.08909}}


\bibitem[\protect\citeauthoryear{Potter and De~Jong}{Potter and
  De~Jong}{1994}]%
        {potter1994cooperative}
\bibfield{author}{\bibinfo{person}{Mitchell~A Potter} {and}
  \bibinfo{person}{Kenneth~A De~Jong}.} \bibinfo{year}{1994}\natexlab{}.
\newblock \showarticletitle{A cooperative coevolutionary approach to function
  optimization}. In \bibinfo{booktitle}{{\em International Conference on
  Parallel Problem Solving from Nature}}. Springer, \bibinfo{pages}{249--257}.
\newblock


\bibitem[\protect\citeauthoryear{Rajpurkar, Irvin, Zhu, Yang, Mehta, Duan,
  Ding, Bagul, Langlotz, Shpanskaya, et~al\mbox{.}}{Rajpurkar
  et~al\mbox{.}}{2017}]%
        {rajpurkar2017chexnet}
\bibfield{author}{\bibinfo{person}{Pranav Rajpurkar}, \bibinfo{person}{Jeremy
  Irvin}, \bibinfo{person}{Kaylie Zhu}, \bibinfo{person}{Brandon Yang},
  \bibinfo{person}{Hershel Mehta}, \bibinfo{person}{Tony Duan},
  \bibinfo{person}{Daisy Ding}, \bibinfo{person}{Aarti Bagul},
  \bibinfo{person}{Curtis Langlotz}, \bibinfo{person}{Katie Shpanskaya},
  {et~al\mbox{.}}} \bibinfo{year}{2017}\natexlab{}.
\newblock \showarticletitle{Chexnet: Radiologist-level pneumonia detection on
  chest x-rays with deep learning}.
\newblock \bibinfo{journal}{{\em arXiv preprint arXiv:1711.05225\/}}
  (\bibinfo{year}{2017}).
\newblock


\bibitem[\protect\citeauthoryear{Real, Aggarwal, Huang, and Le}{Real
  et~al\mbox{.}}{2018}]%
        {real2018regularized}
\bibfield{author}{\bibinfo{person}{Esteban Real}, \bibinfo{person}{Alok
  Aggarwal}, \bibinfo{person}{Yanping Huang}, {and} \bibinfo{person}{Quoc~V
  Le}.} \bibinfo{year}{2018}\natexlab{}.
\newblock \showarticletitle{Regularized evolution for image classifier
  architecture search}.
\newblock \bibinfo{journal}{{\em arXiv preprint arXiv:1802.01548\/}}
  (\bibinfo{year}{2018}).
\newblock


\bibitem[\protect\citeauthoryear{Real, Moore, Selle, et~al\mbox{.}}{Real
  et~al\mbox{.}}{2017}]%
        {real2017large}
\bibfield{author}{\bibinfo{person}{E. Real}, \bibinfo{person}{S. Moore},
  \bibinfo{person}{A. Selle}, {et~al\mbox{.}}} \bibinfo{year}{2017}\natexlab{}.
\newblock \showarticletitle{Large-scale evolution of image classifiers}.
\newblock \bibinfo{journal}{{\em arXiv preprint arXiv:1703.01041\/}}
  (\bibinfo{year}{2017}).
\newblock


\bibitem[\protect\citeauthoryear{Snoek, Larochelle, and Adams}{Snoek
  et~al\mbox{.}}{2012}]%
        {snoek2012practical}
\bibfield{author}{\bibinfo{person}{Jasper Snoek}, \bibinfo{person}{Hugo
  Larochelle}, {and} \bibinfo{person}{Ryan~P Adams}.}
  \bibinfo{year}{2012}\natexlab{}.
\newblock \showarticletitle{Practical bayesian optimization of machine learning
  algorithms}. In \bibinfo{booktitle}{{\em Advances in neural information
  processing systems}}. \bibinfo{pages}{2951--2959}.
\newblock


\bibitem[\protect\citeauthoryear{Snoek, Rippel, Swersky, Kiros, Satish,
  Sundaram, Patwary, Prabhat, and Adams}{Snoek et~al\mbox{.}}{2015}]%
        {snoek2015scalable}
\bibfield{author}{\bibinfo{person}{Jasper Snoek}, \bibinfo{person}{Oren
  Rippel}, \bibinfo{person}{Kevin Swersky}, \bibinfo{person}{Ryan Kiros},
  \bibinfo{person}{Nadathur Satish}, \bibinfo{person}{Narayanan Sundaram},
  \bibinfo{person}{Md~Mostofa~Ali Patwary}, \bibinfo{person}{Mr Prabhat}, {and}
  \bibinfo{person}{Ryan~P Adams}.} \bibinfo{year}{2015}\natexlab{}.
\newblock \showarticletitle{Scalable Bayesian Optimization Using Deep Neural
  Networks.}. In \bibinfo{booktitle}{{\em ICML}}. \bibinfo{pages}{2171--2180}.
\newblock


\bibitem[\protect\citeauthoryear{Stanley and Miikkulainen}{Stanley and
  Miikkulainen}{2002}]%
        {stanley:ec02}
\bibfield{author}{\bibinfo{person}{Kenneth~O. Stanley} {and}
  \bibinfo{person}{Risto Miikkulainen}.} \bibinfo{year}{2002}\natexlab{}.
\newblock \showarticletitle{{E}volving {N}eural {N}etworks {T}hrough
  {A}ugmenting {T}opologies}.
\newblock \bibinfo{journal}{{\em Evolutionary Computation\/}}
  \bibinfo{volume}{10} (\bibinfo{year}{2002}), \bibinfo{pages}{99--127}.
\newblock
\showURL{%
\url{stanley:ec02}}


\bibitem[\protect\citeauthoryear{Suganuma, Shirakawa, and Nagao}{Suganuma
  et~al\mbox{.}}{2017}]%
        {suganuma2017genetic}
\bibfield{author}{\bibinfo{person}{M. Suganuma}, \bibinfo{person}{S.
  Shirakawa}, {and} \bibinfo{person}{T. Nagao}.}
  \bibinfo{year}{2017}\natexlab{}.
\newblock \showarticletitle{A genetic programming approach to designing
  convolutional neural network architectures}. In \bibinfo{booktitle}{{\em
  Proc. of GECCO}}. ACM, \bibinfo{pages}{497--504}.
\newblock


\bibitem[\protect\citeauthoryear{Szegedy, Liu, Jia, Sermanet, Reed, Anguelov,
  Erhan, Vanhoucke, and Rabinovich}{Szegedy et~al\mbox{.}}{2015}]%
        {szegedy2015going}
\bibfield{author}{\bibinfo{person}{Christian Szegedy}, \bibinfo{person}{Wei
  Liu}, \bibinfo{person}{Yangqing Jia}, \bibinfo{person}{Pierre Sermanet},
  \bibinfo{person}{Scott Reed}, \bibinfo{person}{Dragomir Anguelov},
  \bibinfo{person}{Dumitru Erhan}, \bibinfo{person}{Vincent Vanhoucke}, {and}
  \bibinfo{person}{Andrew Rabinovich}.} \bibinfo{year}{2015}\natexlab{}.
\newblock \showarticletitle{Going deeper with convolutions}. In
  \bibinfo{booktitle}{{\em Proceedings of the IEEE Conference on Computer
  Vision and Pattern Recognition}}. \bibinfo{pages}{1--9}.
\newblock


\bibitem[\protect\citeauthoryear{Vincent, Larochelle, Lajoie, Bengio, and
  Manzagol}{Vincent et~al\mbox{.}}{2010}]%
        {vincent2010stacked}
\bibfield{author}{\bibinfo{person}{Pascal Vincent}, \bibinfo{person}{Hugo
  Larochelle}, \bibinfo{person}{Isabelle Lajoie}, \bibinfo{person}{Yoshua
  Bengio}, {and} \bibinfo{person}{Pierre-Antoine Manzagol}.}
  \bibinfo{year}{2010}\natexlab{}.
\newblock \showarticletitle{Stacked denoising autoencoders: Learning useful
  representations in a deep network with a local denoising criterion}.
\newblock \bibinfo{journal}{{\em Journal of machine learning research\/}}
  \bibinfo{volume}{11}, \bibinfo{number}{Dec} (\bibinfo{year}{2010}),
  \bibinfo{pages}{3371--3408}.
\newblock


\bibitem[\protect\citeauthoryear{Wang, Peng, Lu, Lu, Bagheri, and Summers}{Wang
  et~al\mbox{.}}{2017}]%
        {wang2017chestx}
\bibfield{author}{\bibinfo{person}{Xiaosong Wang}, \bibinfo{person}{Yifan
  Peng}, \bibinfo{person}{Le Lu}, \bibinfo{person}{Zhiyong Lu},
  \bibinfo{person}{Mohammadhadi Bagheri}, {and} \bibinfo{person}{Ronald~M
  Summers}.} \bibinfo{year}{2017}\natexlab{}.
\newblock \showarticletitle{Chestx-ray8: Hospital-scale chest x-ray database
  and benchmarks on weakly-supervised classification and localization of common
  thorax diseases}. In \bibinfo{booktitle}{{\em Computer Vision and Pattern
  Recognition (CVPR), 2017 IEEE Conference on}}. IEEE,
  \bibinfo{pages}{3462--3471}.
\newblock


\bibitem[\protect\citeauthoryear{Williams}{Williams}{1992}]%
        {williams1992simple}
\bibfield{author}{\bibinfo{person}{Ronald~J Williams}.}
  \bibinfo{year}{1992}\natexlab{}.
\newblock \showarticletitle{Simple statistical gradient-following algorithms
  for connectionist reinforcement learning}.
\newblock \bibinfo{journal}{{\em Machine learning\/}} \bibinfo{volume}{8},
  \bibinfo{number}{3-4} (\bibinfo{year}{1992}), \bibinfo{pages}{229--256}.
\newblock


\bibitem[\protect\citeauthoryear{Yanai and Iba}{Yanai and Iba}{2001}]%
        {yanai2001multi}
\bibfield{author}{\bibinfo{person}{Kohsuke Yanai} {and}
  \bibinfo{person}{Hitoshi Iba}.} \bibinfo{year}{2001}\natexlab{}.
\newblock \showarticletitle{Multi-agent robot learning by means of genetic
  programming: Solving an escape problem}. In \bibinfo{booktitle}{{\em
  International Conference on Evolvable Systems}}. Springer,
  \bibinfo{pages}{192--203}.
\newblock


\bibitem[\protect\citeauthoryear{Yong and Miikkulainen}{Yong and
  Miikkulainen}{2001}]%
        {yong2001cooperative}
\bibfield{author}{\bibinfo{person}{Chern~Han Yong} {and} \bibinfo{person}{Risto
  Miikkulainen}.} \bibinfo{year}{2001}\natexlab{}.
\newblock \showarticletitle{Cooperative coevolution of multi-agent systems}.
\newblock \bibinfo{journal}{{\em University of Texas at Austin, Austin, TX\/}}
  (\bibinfo{year}{2001}).
\newblock


\bibitem[\protect\citeauthoryear{Zhou, Qu, Li, Zhao, Suganthan, and Zhang}{Zhou
  et~al\mbox{.}}{2011}]%
        {zhou2011multiobjective}
\bibfield{author}{\bibinfo{person}{Aimin Zhou}, \bibinfo{person}{Bo-Yang Qu},
  \bibinfo{person}{Hui Li}, \bibinfo{person}{Shi-Zheng Zhao},
  \bibinfo{person}{Ponnuthurai~Nagaratnam Suganthan}, {and}
  \bibinfo{person}{Qingfu Zhang}.} \bibinfo{year}{2011}\natexlab{}.
\newblock \showarticletitle{Multiobjective evolutionary algorithms: A survey of
  the state of the art}.
\newblock \bibinfo{journal}{{\em Swarm and Evolutionary Computation\/}}
  \bibinfo{volume}{1}, \bibinfo{number}{1} (\bibinfo{year}{2011}),
  \bibinfo{pages}{32--49}.
\newblock


\bibitem[\protect\citeauthoryear{Zoph and Le}{Zoph and Le}{2016}]%
        {zoph:arxiv16}
\bibfield{author}{\bibinfo{person}{Barret Zoph} {and} \bibinfo{person}{Quoc~V.
  Le}.} \bibinfo{year}{2016}\natexlab{}.
\newblock \showarticletitle{Neural Architecture Search with Reinforcement
  Learning}.
\newblock \bibinfo{journal}{{\em CoRR\/}}  \bibinfo{volume}{abs/1611.01578}
  (\bibinfo{year}{2016}).
\newblock
\showURL{%
\url{http://arxiv.org/abs/1611.01578}}


\end{thebibliography}

\end{document}